# Long-term simulation of physical and mechanical behaviors using curriculum-transfer-learning based physics-informed neural networks

Yuan Guo[a], Zhuojia Fu[a,b,*], Jian Min[a], Shiyu Lin[a], Xiaoting Liu[c], Youssef F. Rashed[d], Xiaoying Zhuang[e,*]

[a] *College of Mechanics and Engineering Science, Hohai University, Nanjing 211100, China*
[b] *Key Laboratory of Ministry of Education for Coastal Disaster and Protection, Hohai University, Nanjing 210098, China*
[c] *Institute of Science and Technology Research, China Three Gorges Corporation, Beijing 101199, China*
[d] *Department of Structural Engineering, Cairo University, Giza, Egypt*
[e] *Institute of Photonics, Department of Mathematics and Physics, Leibniz Universität Hannover, Germany*

**Abstract:** This paper proposes a Curriculum-Transfer-Learning based physics-informed neural network (CTL-PINN) for long-term simulation of physical and mechanical behaviors. The main innovation of CTL-PINN lies in decomposing long-term problems into a sequence of short-term sub-problems. Initially, the standard PINN is employed to solve the first sub-problem. As the simulation progresses, subsequent time-domain problems are addressed using a curriculum learning approach that integrates information from previous steps. Furthermore, transfer learning techniques are incorporated, allowing the model to effectively utilize prior training data and solve sequential time-domain transfer problems. CTL-PINN combines the strengths of curriculum learning and transfer learning, overcoming the limitations of standard PINNs, such as local optimization issues, and addressing the inaccuracies over extended time domains encountered in CL-PINN and the low computational efficiency of TL-PINN. The efficacy and robustness of CTL-PINN are demonstrated through applications to nonlinear wave propagation, Kirchhoff plate dynamic response, and the hydrodynamic model of the Three Gorges Reservoir Area, showcasing its superior capability in addressing long-term computational challenges.

## 1 Introduction

Partial differential equations (PDEs) play a crucial role in modeling various natural phenomena, such as time-dependent processes in fluid dynamics, electromagnetism, and thermodynamics. To solve these PDEs, a variety of numerical methods have been developed, including the finite element methods (Zienkiewicz et al., 1977), mesh-free methods (Rabczuk et al., 2019), finite difference methods (Thomas, 2013), and finite integration methods (Zhao et al., 2022), among others. However, traditional numerical methods face significant challenges related to computational complexity, such as high-quality mesh generation, nonlinear problem handling, and scalability issues. These challenges are exacerbated in high-dimensional problems, where computational costs increase substantially, leading to the well-known "curse of dimensionality" (Hao et al., 2022).

---

* Corresponding author
  E-mail address: paul212063@hhu.edu.cn (Z. Fu), zhuang@iop.uni-hannover.de (X. Zhuang).

The rapid development of computational technology has provided new opportunities for scientific research, with machine learning emerging as a powerful tool for addressing complex mathematical models in physics. Numerous studies have applied machine learning techniques to solve PDEs (Beck et al., 2021; Han et al., 2018; Sirignano & Spiliopoulos, 2018), but traditional machine learning algorithms typically rely on large datasets and operate as data-driven approaches. While effective in some cases, these methods often fail to satisfy physical constraints, particularly in real-world engineering and experimental scenarios where data is sparse and noisy. As a result, data-driven methods can introduce significant generalization errors and may diverge from established physical principles. Incorporating known physical laws into machine learning models has thus become a critical area of research, enabling more efficient and robust scientific computation (Karniadakis et al., 2021; C. Meng et al., 2022; Thuerey et al., 2021). Recently, Physics-Informed Neural Networks (Fu et al., 2024; Jeong et al., 2023, 2025; Qian et al., 2024; Raissi et al., 2019; Xiao et al., 2025), which incorporate physical constraints directly into the learning process, have garnered significant attention across various fields. PINN leverage the automatic differentiation capabilities, of neural networks to embed PDEs into the loss function, facilitating a hybrid approach that combines data and physical information to yield solutions that conform to both. This approach effectively reduces the data requirements typical of traditional neural networks, making it particularly suitable for scientific domains with limited data but well-established physical knowledge (Berardi et al., 2025; E. Zhang et al., 2022). However, for time-space-dependent PDEs in large spatiotemporal domains, training PINN remains challenging due to complex structures and high-dimensional input-output relationships. Additionally, for long-duration problems, the increasing solution space and limited intermediate data make neural networks difficult to optimize, resulting in suboptimal performance.

To address the challenges associated with solving PDEs over large spatiotemporal domains and long time horizons, several advanced methodologies have been proposed. The Conservative Physics-Informed Neural Networks (CPINN) applies domain decomposition to nonlinear conservation laws within discrete domains (Jagtap et al., 2020). This approach was later extended to a broader class of problems through the development of Extended Physics-Informed Neural Networks (XPINN) (Jagtap & Karniadakis, 2020). A parallel framework based on CPINN and XPINN was further developed (Shukla et al., 2021). While this approach achieves computational acceleration, it fails to enhance solution accuracy. A hybrid approach combining a coarse global solver with fine PINN operating on parallel subdomains was introduced (X. Meng et al., 2020), where the coarse solver provides approximate solutions that are refined by fine PINN to expedite transient PDE solutions for long-duration problems. Slicing the time domain into multiple segments was proposed (Krishnapriyan et al., 2021), transforming a long-duration problem into a series of short-duration problems. By using the final prediction of each segment as the initial condition for the next, this approach advances the solution iteratively, improving model accuracy. Similarly, A time-domain stacking approach was introduced (Penwarden et al., 2023), allowing partial overlap between time domains to facilitate transfer learning, ensuring continuity across overlapping regions. Pre-training Physics-Informed Neural Networks (PT-PINN) gradually expands the solution domain by breaking down long-duration problems into smaller, more manageable subproblems (Guo et al., 2023). By combining resampling and optimization strategies, PT-PINN enhances accuracy; this time-domain decomposition strategy can be viewed as a form of curriculum learning. A unique multi-step asymptotic pretraining method based on PINN was applied to effectively solve singular perturbation parabolic problems with steep gradients over spatiotemporal domains (F. Cao et al., 2024). An advanced time-stepping PINN (AT-PINN) was developed for time-dependent oscillatory problems (Chen et al., 2024), leveraging techniques such as spatiotemporal normalization, reactivation optimization, transfer learning, and sinusoidal activation functions to improve the accuracy of oscillation simulations.

This study aims to address the problem of solving time-dependent PDEs over extended durations by introducing a generalized approach that enhances PINN accuracy and robustness for this class of equations. To validate and compare the effectiveness of the proposed method, we include several case studies. Our approach employs a time-domain decomposition strategy, segmenting long-duration problems into short-duration intervals. Through the combination of curriculum learning (Bengio et al., 2009) and transfer learning (Weiss et al., 2016), we progressively solve the entire domain. Success in solving these short-duration subproblems lays a foundation for tackling extended duration problems. Based on this framework, we utilize curriculum learning to expand the temporal domain solvable by individual models, establishing a basis for transfer learning in subsequent steps. The main contributions of this paper are as follows:

(1) The proposed CTL-PINN represents an advancement and extension of previous work. The curriculum learning aspect allows a single neural network to solve for the largest possible time domain while maintaining accuracy, thus enabling subsequent transfer learning to achieve larger step sizes. Essentially, the follow-up transfer learning operates as a time-domain stacking decomposition method, further expanding the solvable time domain.

(2) This research focuses more on the study of second-order time derivatives and systems of partial differential equations (PDEs), supplementing previous work in solving time-dependent PDEs. The CTL-PINN method is applied to inverse problems, including the inversion of source terms and time-varying parameters in time-dependent PDEs.

(3) For the first time, physics-informed machine learning technique is applied to the Three Gorges Reservoir Area, achieving inversion of roughness parameters and prediction of flow and water levels based on limited measured data and a hydrodynamic model with unknown roughness parameters.

A brief outline of the paper is given as follows. Section 2 briefly introduces the concept of standard PINN. Section 3 discusses Curriculum Learning, Transfer Learning, and the implementation principles and processes of CTL-PINN. Section 4 examines the feasibility and accuracy of the CTL-PINN method in solving long-term evolution forward and inverse problems through examples such as nonlinear wave propagation, dynamic response of thin plates, and the hydrodynamic model of the Three Gorges Reservoir Area. Finally, Section 5 summarizes the conclusions drawn from this study.

## 2 Physics-informed neural networks

In this paper we focus on the time dependent PDEs of the form:

$$\frac{\partial^k u}{\partial t^k} + \mathcal{N}[u] = 0 \qquad (X,t) \in \Omega \times (0,T] (k=1,2) \tag{1}$$

where $X$ represents the spatial position, $t$ represents time, $\Omega$ is the spatial domain, $T$ is the final time for solving, $u = u(X,t)$ is the function to be solved, and $\mathcal{N}$ represents a partial differential operator with respect to the spatial variable. Taking $k = 2$ as an example, its initial condition is:

$$\begin{cases} u(X,0) = \phi(X) & X \in \Omega \\ \dfrac{\partial u(X,0)}{\partial t} = \varphi(X) & X \in \Omega \end{cases} \tag{2}$$

where $\phi(X)$ and $\varphi(X)$ are given functions representing two types of initial conditions. The boundary conditions are:

$$\begin{cases} u(X,t) = \bar{u}(X,t) & (X,t) \in \Gamma_1 \times (0,T] \\ \dfrac{\partial u(X,t)}{\partial n} = \bar{q}(X,t) & (X,t) \in \Gamma_2 \times (0,T] \end{cases} \tag{3}$$

where $\Gamma_1 \cup \Gamma_2 = \Gamma$, $\Gamma_1 \cap \Gamma_2 = \varnothing$, $\Gamma = \partial \Omega$ represents the boundary of $\Omega$, $\Gamma_1$ represents the first type boundary condition or Dirichlet condition, $\Gamma_2$ represents the second type boundary condition or Neumann condition, $n$ is the unit outward normal vector on $\Gamma_2$, $\bar{u}(X,t)$ and $\bar{q}(X,t)$ are given functions.

The standard PINN method approximates the solution of equation (1) using a Deep Neural Network. As shown in Figure 1, the inputs of the neural network are the spatiotemporal coordinates $(X,t)$, corresponding to the spatiotemporal variables of equation (1). The output is the approximate solution of equation (1), denoted as $u_\theta$, where $\theta$ represents the parameters to be optimized in the neural network. The loss function reflects the physical information of the equation, comprising three parts: the residual terms of the initial and boundary conditions, and the residual term of the governing equation, as shown below:

$$\mathcal{L}(\theta; \tau_i, \tau_b, \tau_r) = w_i \mathcal{L}_i(\theta; \tau_i) + w_b \mathcal{L}_b(\theta; \tau_b) + w_r \mathcal{L}_r(\theta; \tau_r) \tag{4}$$

$$\mathcal{L}_i(\theta; \tau_i) = \frac{1}{N_i} \left( \sum_{i=1}^{N_i^1} \left| u_\theta(X_i, 0) - \phi(X_i) \right|^2 + \sum_{i=1}^{N_i^2} \left| \frac{\partial u_\theta(X_i, 0)}{\partial t} - \varphi(X_i) \right|^2 \right) \tag{5}$$

$$\mathcal{L}_b(\theta; \tau_b) = \frac{1}{N_b} \left( \sum_{i=1}^{N_b^1} \left| u_\theta(X_i, t_i) - \bar{u}(X_i, t_i) \right|^2 + \sum_{i=1}^{N_b^2} \left| \frac{\partial u_\theta(X_i, t_i)}{\partial n} - \bar{q}(X_i, t_i) \right|^2 \right) \tag{6}$$

$$\mathcal{L}_r(\theta; \tau_r) = \frac{1}{N_r} \sum_{i=1}^{N_r} \left( \frac{\partial^2 u_\theta(X_i, t_i)}{\partial t^2} + \mathcal{N}[u_\theta(X_i, t_i)] \right)^2 \tag{7}$$

In equation (4), $w_i$, $w_b$, and $w_r$ are the weights of the three parts of the loss function, $\mathcal{L}_i$ and $\mathcal{L}_b$ represent the supervised learning losses on the initial and boundary conditions, and $\mathcal{L}_r$ represents the residual loss defined on the residual dataset. In equations (5), (6), and (7), $u_\theta(X,t)$ is the neural network prediction, with its various derivatives obtained through Automatic Differentiation. Based on the two types of initial conditions for velocity and displacement, and the two types of boundary conditions, $N_i$ and $N_b$ are divided into two parts, $N_i = N_i^1 + N_i^2$, $N_b = N_b^1 + N_b^2$, and $N_r$ denotes the number of training points, which are obtained through sampling. The datasets of initial and boundary conditions, as well as the residual dataset, are saved as:

$$\tau_i = \{(X_j, 0, \phi(X_j))| X_j \in \Omega\}_{j=1}^{N_i^1} \cup \{(X_j, 0)| X_j \in \Omega\}_{j=1}^{N_i^2} \tag{8}$$

$$\tau_b = \{(X_i, t_i, \bar{u}(X_i, t_i))|(X_i, t_i) \in \Gamma_1 \times (0,T]\}_{i=1}^{N_b^1} \cup \{(X_i, t_i)|(X_i, t_i) \in \Gamma_2 \times (0,T]\}_{i=1}^{N_b^2} \tag{9}$$

$$\tau_r = \{(X_i, t_i) \in \Omega \times (0,T]\}_{i=1}^{N_r} \tag{10}$$

By minimizing the loss function, the approximate solution $u_\theta$ output by the neural network approaches the true solution. PINN can obtain predictions that satisfy the physical information within the solving domain. Let $\Sigma$ denote the sum of the datasets, i.e., $\Sigma = \{\tau_i, \tau_b, \tau_r\}$, the training of PINN is an optimization process to minimize the loss function:

$$\bar{\theta} = \arg\min_\theta \mathcal{L}(\theta; \Sigma) \tag{11}$$

Commonly used optimization algorithms for neural networks include the Adam algorithm and the L-BFGS algorithm. The update of the neural network parameters $\theta$ is achieved through the backpropagation algorithm. The activation function $\sigma$ is higher-order differentiable, with commonly used functions being tanh and sin.

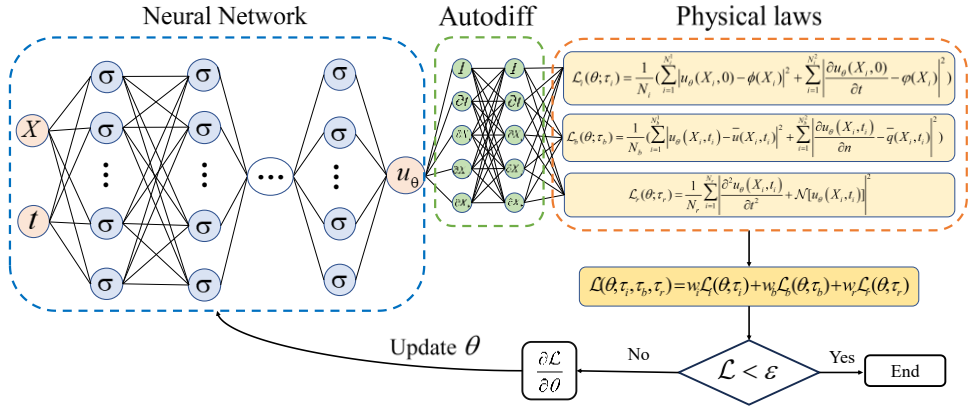

Fig 1   Solving time dependent PDEs with standard PINN

## 3   Curriculum-Transfer-Learning based physics-informed neural networks

This study introduces curriculum learning and transfer learning techniques, proposes the Curriculum-Transfer-Learning based physics-informed neural network (CTL-PINN), and applies it to solving long-term evolution problems. As shown in Figure 2, the first step of CTL-PINN in solving long-term problems is to obtain a model trained by the standard PINN method over a short time. The standard PINN method can be effectively trained to solve short-term problems, assuming a time domain range of $(0, T_p]$. Based on this, curriculum learning starts, expanding the time domain with a step size of $\Delta T_{pc}$, extending the solvable time domain to $(0, T_c]$. After completing curriculum learning, transfer learning begins, using a step size of $\Delta T_{ct}$ to perform time-domain transfer, ultimately extending the solvable time domain to $(0, T_c]$. The specific implementation steps of CTL-PINN can be found in Algorithm 1, and the principles of curriculum learning and transfer learning are detailed in Sections 3.1 and 3.2.

Curriculum learning allows a single neural network to solve for the largest possible time domain while maintaining accuracy, enabling subsequent transfer learning to achieve larger step sizes. Essentially, the subsequent transfer learning acts as a time-domain stacking decomposition method, further expanding the solvable time domain. The specific operations performed in a certain training step by curriculum learning and transfer learning are shown in Figure 3, where each training step can be regarded as an independent PINN solving process.

---

Algorithm 1: Steps for solving long-term evolution problems using Curriculum-Transfer-Learning based physics-informed neural network (CTL-PINN)

Input:
  Neural network structure;
  Curriculum learning intervals: $(0, T_p], \cdots (0, T_p + k\Delta T_{pc}], \cdots (0, T_c]$
  Transfer learning intervals: $(0, T_c], \cdots (k\Delta T_{ct}, T_c + k\Delta T_{ct}], \cdots (mT_{ct}, T_t]$

Curriculum learning steps:
  Initialize the neural network
  Solve the equation in the time domain $(0, T_p]$ using the standard PINN method, and save the neural network parameters $\theta_{pc}^0$
  for $k = 1, 2, \cdots, n$ do:
    Generate the additional supervised learning dataset $\tau_{sp}^{(k)}$ and the training dataset $\Sigma_{pc}^{(k)}$
    Minimize the loss function $\theta_{pc}^{(k)} = \arg\min_\theta \mathcal{L}(\theta; \Sigma_{pc}^{(k)}, \theta_{pc}^{(k-1)})$
    Save the neural network parameters $\theta_{pc}^{(k)}$ and use them as the initialization parameters for the next model
  end

Transfer learning steps:
  $\theta_{ct}^{(0)} = \theta_{pc}^{(n)}$ as the initialization parameters
  for $k = 1, 2, \cdots, m$ do:
    Generate the additional supervised learning dataset $\tau_{sp}^{(k)}$, the initial condition dataset $\tau_{ict}^{(k)}$, and the training dataset $\Sigma_{ct}^{(k)}$
    Minimize the loss function $\theta_{ct}^{(k)} = \arg\min_\theta \mathcal{L}(\theta; \Sigma_{ct}^{(k)}, \theta_{ct}^{(k-1)})$
    Save the neural network parameters $\theta_{ct}^{(k)}$ and use them as the initialization parameters for the next model
  end
  The solution to the equation over the long time domain is obtained from $\theta_{ct}^{(1)}, \cdots \theta_{ct}^{(k)}, \cdots \theta_{ct}^{(m)}$

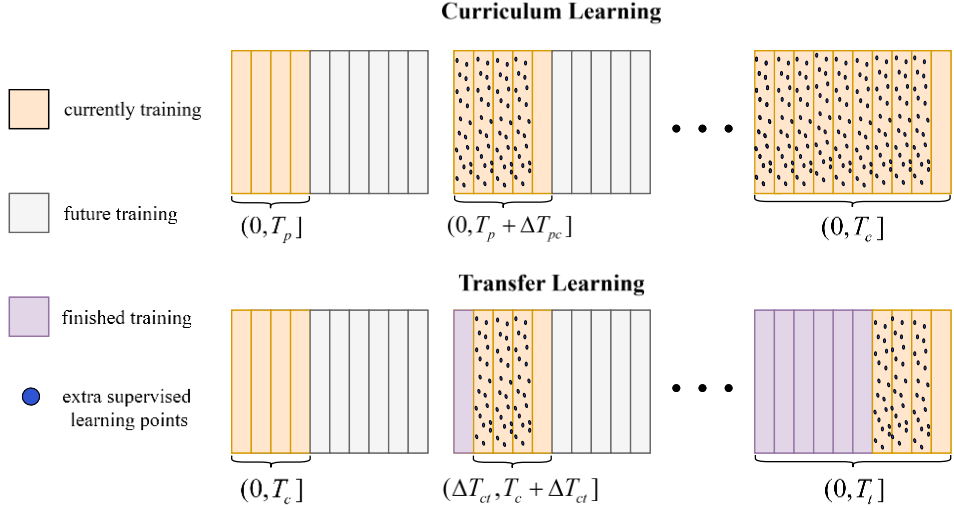

Fig 2　Schematic diagram of Curriculum-Transfer-Learning-based physics-informed neural networks solving long-term problem

### 3.1　Curriculum Learning

Curriculum learning is a training strategy that mimics the human learning process. It suggests that the model should start learning from easy samples and gradually progress to more complex samples and knowledge (Bengio et al., 2009). In this study, the standard PINN is used to successfully complete the training of models over smaller time domains, progressively extending the time domain. If the curriculum learning phase undergoes $n$ time domain expansions, it finally extends to the time domain $(0, T_c]$. The expansion process is described as:

$$(0,T_p],\cdots(0,T_p+k\Delta T_{pc}],\cdots[0,T_c](T_c=T_p+n\Delta T_{pc}; k=1,2,3,......n) \quad (12)$$

In a certain training step of curriculum learning, the model of the source time domain $(0, T_p + (k-1)\Delta T_{pc}]$ and the model of the target time domain $(0, T_p + k\Delta T_{pc}]$ share many common characteristics. They have the same governing equations, initial conditions, and some boundary conditions. Therefore, the model parameters $\theta_{pc}^{(k-1)}$ from the trained source time domain can be imported into the target time domain model. Additionally, extra supervised learning points can be extracted from the intersection of the two time domains $(0, T_p + (k-1)\Delta T_{pc}]$ to facilitate training in the target time domain, thereby reducing the training difficulty and improving the performance of the neural network.

Let $u_{\theta_{pc}}^{(k-1)}(X,t)$ denote the prediction of the neural network trained in the time domain $(0, T_p + (k-1)\Delta T_{pc}]$ at $(X,t)$. Let $N_{sp}$ represent the number of additional supervised learning points, and the additional supervised learning dataset $\tau_{sp}^{(k)}$ is shown as follows:

$$\tau_{sp}^{(k)} = \{(X_i, t_i, u_{\theta_{pc}}^{(k-1)}(X_i, t_i)) | (X_i, t_i) \in \Omega \times (0, T_p + (k-1)\Delta T_{pc}]\}_{i=1}^{N_{sp}} \quad (13)$$

Let $\Sigma_{pc}^{(k)} = \{\tau_{ipc}^{(k)}, \tau_{bpc}^{(k)}, \tau_{rpc}^{(k)}, \tau_{sp}^{(k)}\}$ represent the training dataset for the target time domain $(0, T_p + k\Delta T_{pc}]$. Then, the following optimization problem exists in this time domain:

$$\theta_{pc}^{(k)} = \arg\min_{\theta} \mathcal{L}(\theta; \Sigma_{pc}^{(k)}, \theta_{pc}^{(k-1)}) \tag{14}$$

The loss function is shown as follows:

$$\mathcal{L}(\theta; \Sigma_{pc}^{(k)}, \theta_{pc}^{(k-1)}) = \mathcal{L}(\theta; \tau_{ipc}^{(k)}, \tau_{bpc}^{(k)}, \tau_{rpc}^{(k)}) + w_{sp}\mathcal{L}_{sp}(\theta; \tau_{sp}^{(k)}) \tag{15}$$

where $w_{sp}$ is the weight of the additional supervised learning part, and $\mathcal{L}_{sp}$ is the loss of the additional supervised learning part. The expression is as follows:

$$\mathcal{L}_{sp}(\theta; \tau_{sp}^{(k)}) = \frac{1}{N_{sp}} \sum_{i=1}^{N_{sp}} \left| u_\theta(X_i, t_i) - u_{\theta_{pc}^{(k-1)}}(X_i, t_i) \right|^2 \tag{16}$$

During the curriculum learning process, in order to ensure the accuracy of the training results, the step size should not be too large. Each curriculum learning step can expand the solvable time domain, but due to the difficulty of training the neural network over a large time domain and the limited fitting ability of a single neural network, curriculum learning cannot continue indefinitely. Therefore, it is recommended that $\Delta T_{pc} \leq \frac{1}{4} T_p$, and $T_p < T_c \leq 3T_p$.

### 3.2 Transfer Learning

Transfer learning is a machine learning technique that applies a pre-trained model to a new problem (Weiss et al., 2016). In this study, based on the model of the final time domain $(0, T_c]$ obtained from curriculum learning, the solvable domain is extended to $(0, T_t]$. If the transfer learning phase undergoes $m$ time domain transfers, the solvable domain is eventually extended from $(0, T_c]$ to $(0, T_t]$. The transfer process is described as follows:

$$(0, T_c], \cdots (k\Delta T_{ct}, T_c + k\Delta T_{ct}], \cdots [m T_{ct}, T_t](T_t = T_c + m\Delta T_{ct}; k = 1, 2, 3, \ldots m) \tag{17}$$

In a certain training step of transfer learning, the model of the source time domain $((k-1)\Delta T_{ct}, T_c + (k-1)\Delta T_{ct}]$ and the model of the target time domain $(k\Delta T_{ct}, T_c + k\Delta T_{ct}]$ share many common characteristics. They have the same governing equations, the same time domain length for the solution, and some identical boundary conditions. Therefore, the model parameters $\theta_{ct}^{(k-1)}$ from the trained source time domain can be imported into the target time domain model. Additionally, extra supervised learning points can be extracted from the intersection of the two time domains $(k\Delta T_{ct}, T_c + (k-1)\Delta T_{ct}]$ to facilitate training in the target time domain. The initial conditions for the target time domain can be predicted by the source time domain model. By using multiple neural networks and employing a time-domain stacking decomposition method, the solvable time domain can be expanded.

Let $u_{\theta_{ct}^{(k-1)}}(X, t)$ denote the prediction of the neural network trained in the time domain $((k-1)\Delta T_{ct}, T_c + (k-1)\Delta T_{ct}]$ at $(X, t)$. Then, the additional supervised learning dataset $\tau_{sp}^{(k)}$ is shown as follows:

$$\tau_{sp}^{(k)} = \{(X_i, t_i, u_{\theta_{ct}^{(k-1)}}(X_i, t_i)) | (X_i, t_i) \in \Omega \times (k\Delta T_{ct}, T_c + (k-1)\Delta T_{ct}]\}_{i=1}^{N_{sp}} \tag{18}$$

The initial conditions of the target time domain change, and the initial condition dataset $\tau_{ict}^{(k)}$ is

as follows:

$$\tau_{ict}^{(k)} = \{(X_j, \Delta T_{ct}^k, u_{\theta_{ct}}^{(k-1)}(X_j, \Delta T_{ct}^k))| \ X_j \in \Omega\}_{j=1}^{N_i^1}$$
$$\bigcup \{(X_j, \Delta T_{ct}^k, \frac{\partial u_{\theta_{ct}}^{(k-1)}(X_j, \Delta T_{ct}^k)}{\partial t})| \ X_j \in \Omega\}_{j=1}^{N_i^2} \quad (19)$$

Let $\Sigma_{ct}^{(k)} = \{\tau_{ict}^{(k)}, \tau_{bct}^{(k)}, \tau_{rct}^{(k)}, \tau_{sp}^{(k)}\}$ represent the training dataset for the target time domain $(k\Delta T_{ct}, T_c + k\Delta T_{ct}]$. Then, the following optimization problem exists in this time domain:

$$\theta_{ct}^{(k)} = \arg\min_\theta \mathcal{L}(\theta; \Sigma_{ct}^{(k)}, \theta_{ct}^{(k-1)}) \quad (20)$$

The loss function is shown as follows:

$$\mathcal{L}(\theta; \Sigma_{ct}^{(k)}, \theta_{ct}^{(k-1)}) = \mathcal{L}(\theta; \tau_{ict}^{(k)}, \tau_{bct}^{(k)}, \tau_{rct}^{(k)}) + w_{sp}\mathcal{L}_{sp}(\theta; \tau_{sp}^{(k)}) \quad (21)$$

where $w_{sp}$ is the weight of the additional supervised learning part, and $\mathcal{L}_{sp}$ is the loss of the additional supervised learning part. The expression is as follows:

$$\mathcal{L}_{sp}(\theta; \tau_{sp}^{(k)}) = \frac{1}{N_{sp}} \sum_{i=1}^{N_{sp}} \left| u_\theta(X_i, t_i) - u_{\theta_{ct}}^{(k-1)}(X_i, t_i) \right|^2 \quad (22)$$

During the transfer learning process, to ensure the accuracy of the training results, the intersection of the time domains between two adjacent training steps should not be too small, and the step size should not be too large. Therefore, we recommend $\Delta T_{ct} \leq \frac{1}{4} T_c$.

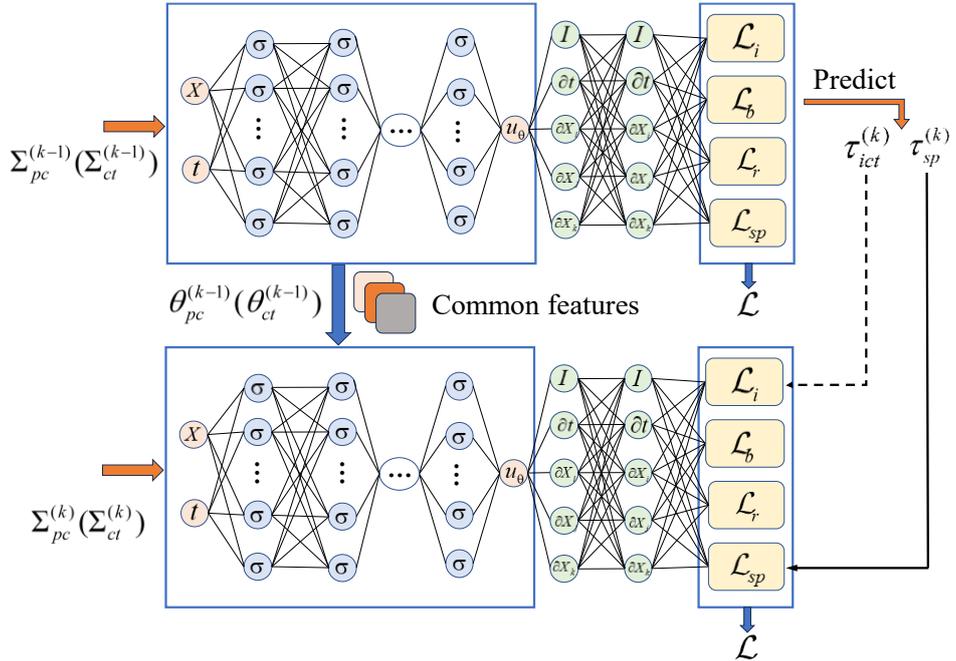

Fig 3　Specific operations performed in a training step of Curriculum learning or transfer learning

## 4 Numerical examples

In this section, the Curriculum-Transfer-Learning based physics-informed neural network (CTL-PINN) is applied to nonlinear wave propagation, Kirchhoff plate dynamic response, and the hydrodynamic model of the Three Gorges Reservoir Area. The effectiveness and robustness of the CTL-PINN are verified through numerical examples. To evaluate the accuracy of the solution, the $L_2$ error is introduced, and its calculation method is as follows:

$$Error = \frac{\sqrt{\sum_{i=1}^{N}|u_\theta(X_i,t_i)-u(X_i,t_i)|^2}}{\sqrt{\sum_{i=1}^{N}|u(X_i,t_i)|^2}} \qquad (23)$$

Where $\{(X_i,t_i)\}_{i=1}^{N}$ represent the test set, $u_\theta(X_i,t_i)$ be the neural network prediction at the test points, and $u(X_i,t_i)$ be the reference values. In Section 4.1, the test set randomly samples $1\times10^4$ points within the entire computational domain. In Section 4.2, the test set uniformly samples $1\times10^3$ points in the time domain at the center of the plate (0.5,0.5). In Section 4.3, the test set comprises specified measured data. The activation function of the neural network in this section is tanh. In Sections 4.1 and 4.2, higher-order derivatives need to be computed. Considering computational efficiency, the L-BFGS algorithm is selected as the optimization algorithm, with specific parameters consistent with those reported in reference (Raissi et al., 2019). In Section 4.3, the Adam algorithm is selected as the optimization algorithm, with the number of iterations set to $6\times10^5$ and the learning rate set to $1\times10^{-3}$.

### 4.1 Nonlinear wave propagation

Nonlinear waves are widely present in numerous natural phenomena and engineering applications (Younas et al., 2024), such as water waves (Zheng et al., 2023), light waves (Roy et al., 2023), sound waves (Quan et al., 2016), seismic waves (Tasaketh et al., 2022), etc. The analysis and simulation of nonlinear wave propagation are of significant importance for understanding and controlling these phenomena. This section considers the following transient nonlinear wave propagation problem, with its governing equation expressed as:

$$\frac{\partial^2 u}{\partial t^2} = c^2 \nabla^2 u + \sin(u) + Q(x,y,z,t) \qquad (x,y,z,t) \in \Omega \times (0,T] \qquad (24)$$

where $c$ is the wave propagation speed, $u=u(x,y,z,t)$ is the displacement to be solved, $Q(x,y,z,t)$ is the wave source function, and $\sin(u)$ is the nonlinear term. The initial conditions are:

$$\begin{cases} u(x,y,z,0) = \cos(x+y+z) & (x,y,z) \in \Omega \\ \dfrac{\partial u(x,y,z,0)}{\partial t} = 0.2\pi\sin(x+y+z) - \dfrac{1}{40}\cos(x+y+z) & (x,y,z) \in \Omega \end{cases} \qquad (25)$$

The boundary conditions are:

$$u(x,y,z,t) = \cos(x+y+z-0.2\pi t)\frac{40}{t+40} \qquad (x,y,z,t) \in \partial\Omega \times (0,T] \qquad (26)$$

The analytical solution is given by $u(x,y,z,t) = \cos(x+y+z-0.2\pi t)\dfrac{40}{t+40}$, and the source term $Q(x,y,z,t)$ can be determined by substituting the analytical solution into the governing equation. The specific mathematical expression is as follows:

$$Q(x,y,z,,t) = -\sin[\cos(x+y+z-0.2\pi t)\frac{40}{t+40}] + (3-0.04\pi^2)\cos(x+y+z-0.2\pi t)\frac{40}{t+40}$$
$$-0.4\pi\sin(x+y+z-0.2\pi t)\frac{40}{(t+40)^2} + \cos(x+y+z-0.2\pi t)\frac{80}{(t+40)^3} \quad (27)$$

In this section, we set $c=1$. As shown in Figure 4, the computational domain $\Omega$ is an annulus with an inner radius of 0.6 and an outer radius of 1. The neural network model has 15 hidden layers, each containing 40 neurons. The weights of the loss functions are all set to 1. The number of initial points is $N_i = 200$, the number of boundary points is $N_b = 500$, and the number of residual points is $N_r = 1200$. These points are obtained using Latin Hypercube Sampling within the corresponding domains. For curriculum learning and transfer learning, the number of additional supervised learning points is $N_{sp} = 1200$.

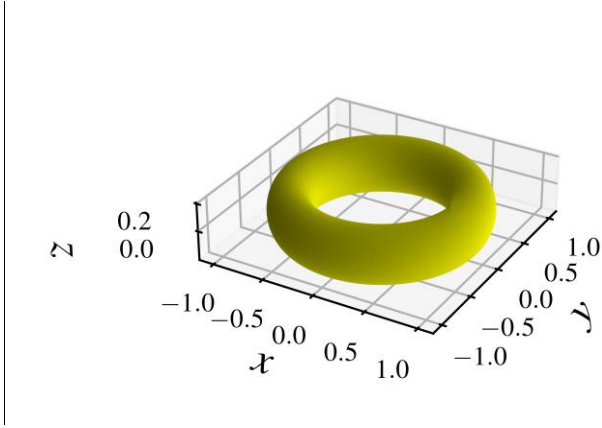

Fig 4  Schematic diagram of circular ring

### 4.1.1 Forward problem

The forward problem is to find an approximate solution that satisfies the governing equation, initial conditions, and boundary conditions. Figure 5 plots the errors for different solution time domains $(0,T]$ using the standard PINN, CT-PINN (only using Curriculum Learning), TL-PINN (only using Transfer Learning), and CTL-PINN. It can be observed that the standard PINN method can achieve high-accuracy results in a short time, which is the foundation for the implementation of CT-PINN, TL-PINN, and CTL-PINN. However, as the required solution time domain increases, the error of the standard PINN solution increases rapidly and becomes unstable, whereas the results trained by CT-PINN, TL-PINN, and CTL-PINN are more accurate and stable.

CL-PINN is based on the model trained by the standard PINN within $(0, 20s]$ and performs curriculum learning with a step size of 5s. When $T \leq 50s$, the error is relatively low. At $T = 80s$, there is a significant accumulation of error. At $T = 220s$, the error becomes large and starts to show instability, and at $T = 250s$, the error increases dramatically. TL-PINN is also based on the model trained by the standard PINN within $(0, 20s]$ and performs transfer learning with a step size of 5s. There is some accumulation of error during the process, but it is smaller than that of CL-PINN. CTL-PINN first performs curriculum learning with a step size of 5s to extend the solution time domain to $(0,50s]$, and then performs transfer learning with a step size of 10s. There is some accumulation of error during the process, but it is smaller than that of both CL-PINN and TL-PINN. The accumulation of error during the propagation process of CT-PINN, TL-PINN, and CTL-PINN is because the model prediction for the additional supervised learning points in the previous step has some errors. In the propagation process of CTL-PINN, 6 steps of curriculum learning and 20 steps of transfer learning are used, totaling 26 steps, whereas CT-PINN and TL-PINN respectively use 46 steps of curriculum learning and transfer learning. CTL-PINN achieves more accurate and stable results with fewer training steps and larger learning steps.

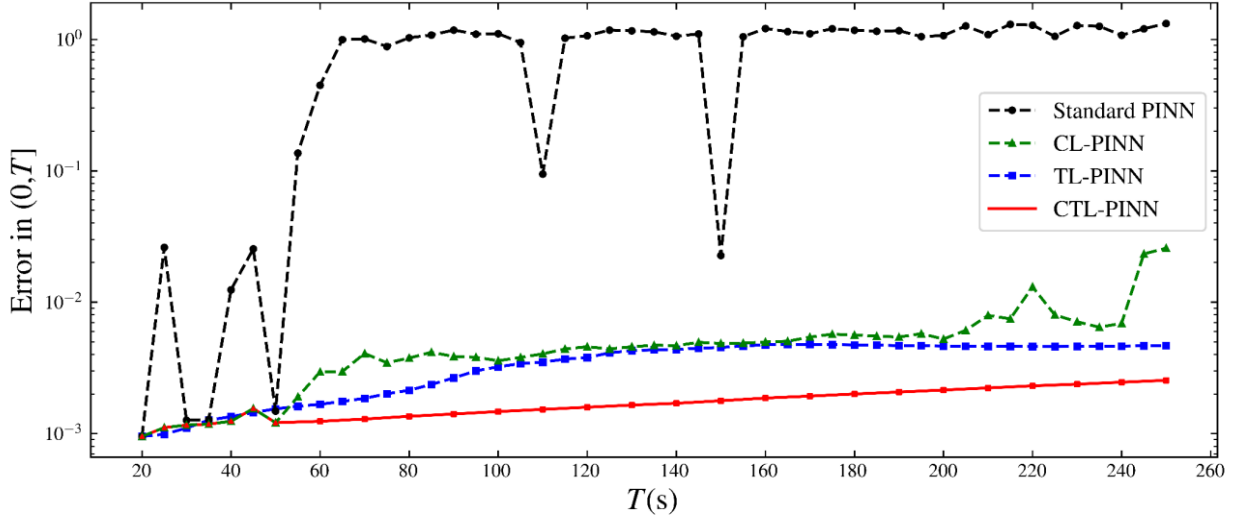

Fig 5　Error variation with respect to the increase of the time domain by standard PINN，CT-PINN，TL-PINN and CTL-PINN

Figure 6 illustrates the predicted solution and the exact solution at the point $(1,0,0)$ over the time domain $(0, 250s]$ for standard PINN and CTL-PINN when $T = 250s$. It can be observed that the CTL-PINN predicted solution is in close agreement with the analytical solution, whereas the standard PINN predicted solution significantly deviates from the analytical solution. This demonstrates the effectiveness of CTL-PINN in simulating long-term nonlinear wave propagation.

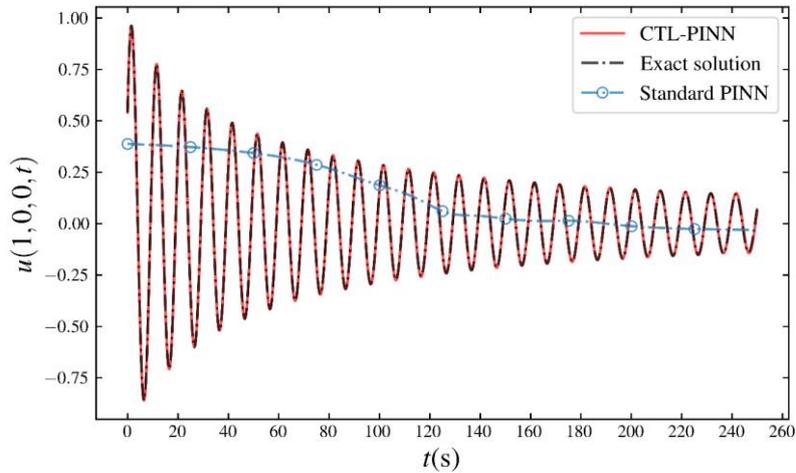

Fig 6  Exact solution, predicted solution by standard PINN and CTL-PINN at point (1, 0, 0)

Figure 7 depicts the distributions of the Exact solution, CTL-PINN, standard PINN, and CTL-PINN Absolute error at $t = 250\text{s}$. The difference between CTL-PINN and the Exact solution is small, with an absolute error within $3 \times 10^{-3}$, while the standard PINN shows significant deviation from the analytical solution. This demonstrates the effectiveness of CTL-PINN in simulating long-term nonlinear wave propagation.

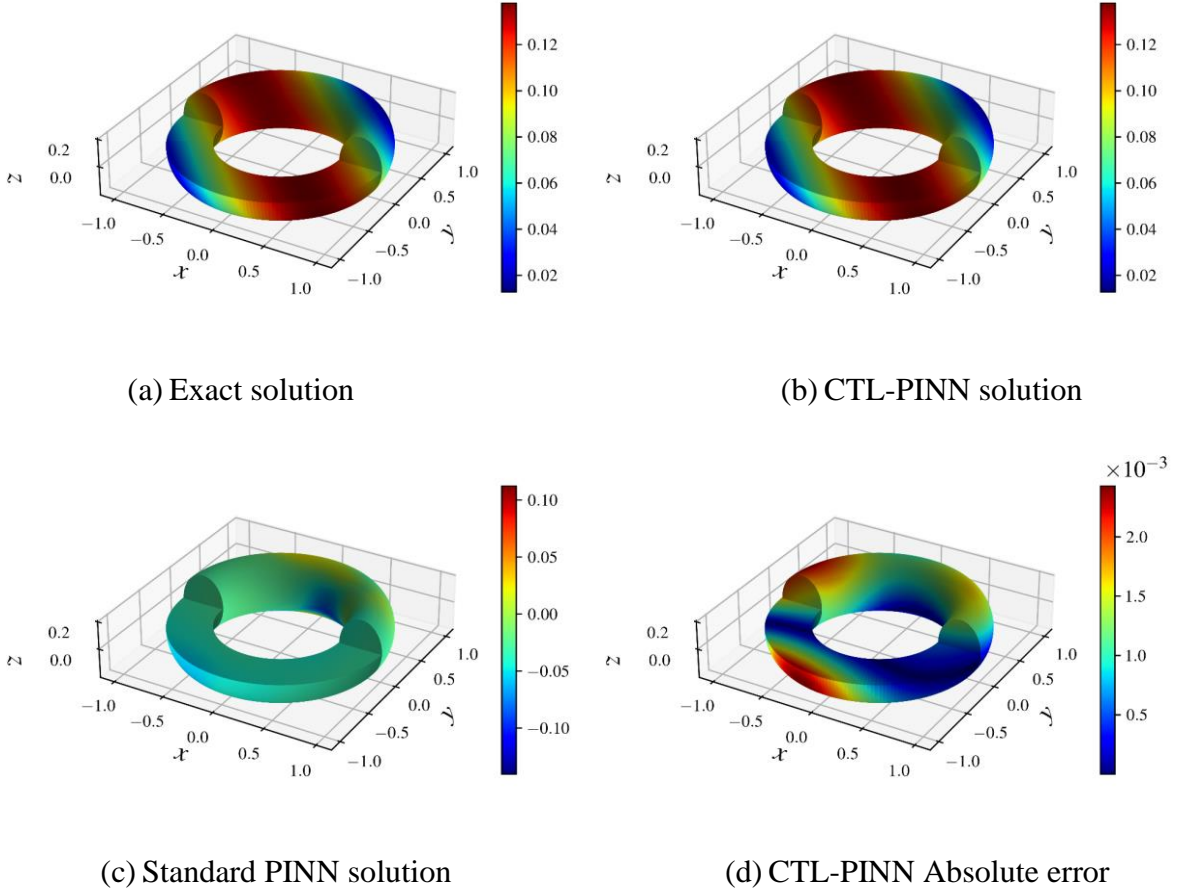

(a) Exact solution  (b) CTL-PINN solution

(c) Standard PINN solution  (d) CTL-PINN Absolute error

Fig 7  At $t = 250s$, the distributions of u (a) Exact solution; (b) CTL-PINN solution; (c) Standard PINN solution; (d) CTL-PINN Absolute error

#### 4.1.2 Inverse problem

In this section, the inverse problem is set as a source term inversion problem, where the initial conditions and boundary conditions are known, but the source term function in the governing equation is unknown. The source term function $Q(x, y, z, t)$ is a function of time. To invert the source term, we need data from several measurement points within the computational domain $\Omega$. In this section, we select the points $(0.8, 0, 0)$, $(-0.8, 0, 0)$, $(0, 0.8, 0)$, and $(0, -0.8, 0)$ as measurement points, assuming that data is measured at these points every second. We transform the output of the neural network into the approximate solution of the equation $u_\theta$ and the approximate solution of the source term function $Q_\theta$. The source term $Q(x, y, z, t)$ in the residual loss $\mathcal{L}_r$ is replaced by the approximate solution $Q_\theta$.

The loss function additionally incorporates the supervised learning loss from the measurement data:

$$\mathcal{L}_l(\theta;\tau_l) = \frac{1}{N_l}\sum_{i=1}^{N_l}|u_\theta(X_i,t_i) - u_l(X_i,t_i)|^2 \qquad (28)$$

where $N_l$ is the number of measurement data points, and $\tau_l$ is the set of measurement data. Through such simple modifications, the curriculum-transfer learning method can be conveniently integrated to invert the long-term source term function.

Based on the model trained by the standard PINN within $(0, 20s]$, we perform curriculum learning in 4 steps with a step size of 5s. Then, based on the model from curriculum learning, we perform transfer learning in 16 steps with a step size of 10s, ultimately extending the solvable time domain to $(0, 200s]$. Figures 8(a) and 8(b) respectively plot the changes in the errors of $u$ and $Q$ as the time domain increases. It can be observed that, compared to the standard PINN method, the CTL-PINN method produces more accurate and stable training results, extending the solvable time domain. Even for the more challenging scenario where the measurement data contains 1% random noise, CTL-PINN achieves good results, demonstrating the robustness of the CTL-PINN method. An interesting finding is that in the inverse problem, using data from measurement points within the computational domain during the propagation process, the error accumulation is not as severe as in the forward problem, indicating that the data from measurement points can somewhat correct the training results.

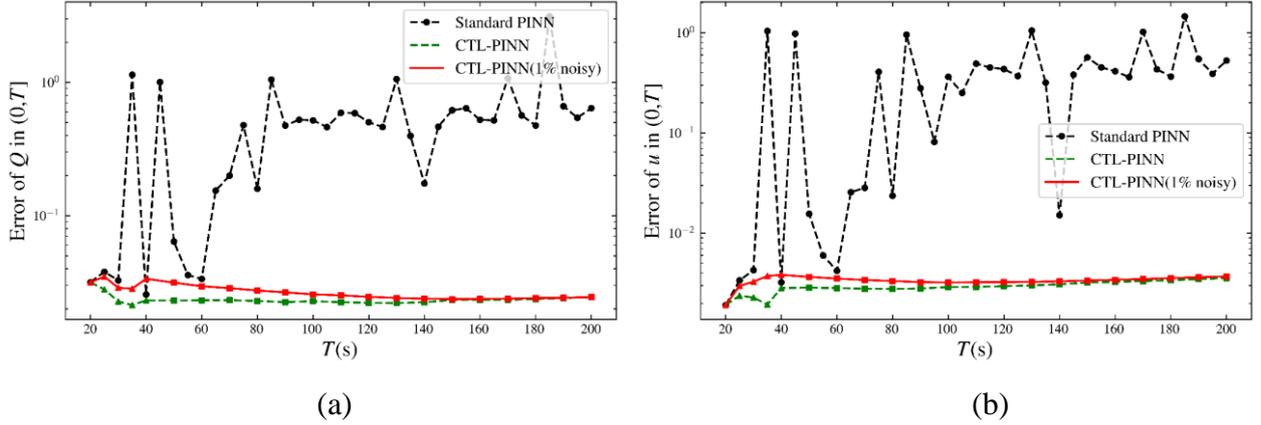

(a)          (b)

Fig 8    Error of (a) $u$ and (b) $Q$ variation with respect to the increase of the time domain in the inverse problem by standard PINN and CTL-PINN

Figure 9 illustrates the source term function inverted by the standard PINN method and the CTL-PINN method, as well as the exact source term function at the point $(1,0,0)$ over the time domain $(0, 200s]$ when $T = 200s$. It can be observed that the source term function inverted by CTL-PINN is in close agreement with the exact solution, whereas the source term function inverted by the standard PINN shows significant deviation from the exact solution. This demonstrates that CTL-PINN can be used to invert the long-term source term of nonlinear waves.

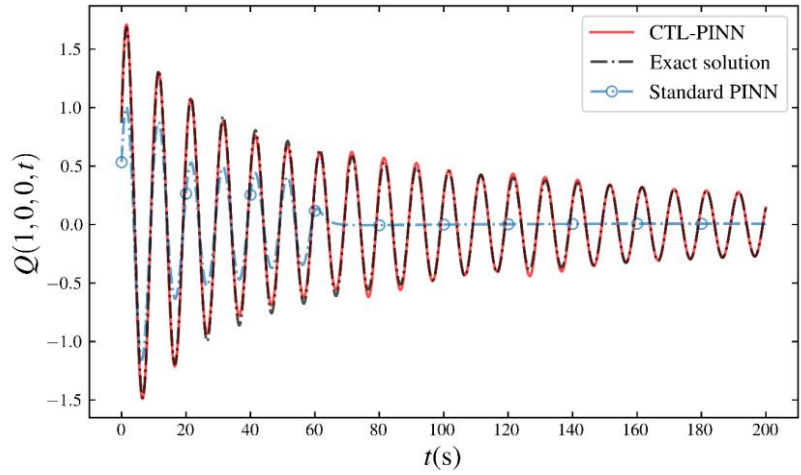

Fig 9  Exact solutions and source term functions obtained from standard PINN and CTL-PINN inversion at point (1, 0, 0)

Figure 10 depicts the distributions of the Exact solution, CTL-PINN, standard PINN, and CTL-PINN Absolute error at $t=200\text{s}$. The difference between the CTL-PINN and the Exact solution is small, with an absolute error within $1.5\times10^{-2}$, while the standard PINN shows significant deviation from the Exact solution. This demonstrates the effectiveness of CTL-PINN in inverting the source term of long-term nonlinear waves.

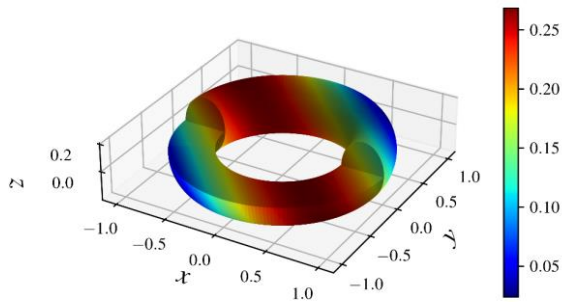

(a) Exact solution

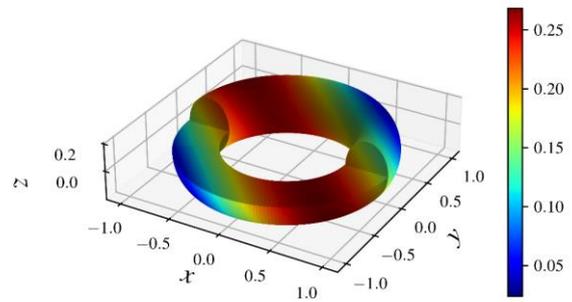

(b) CTL-PINN inversion solution

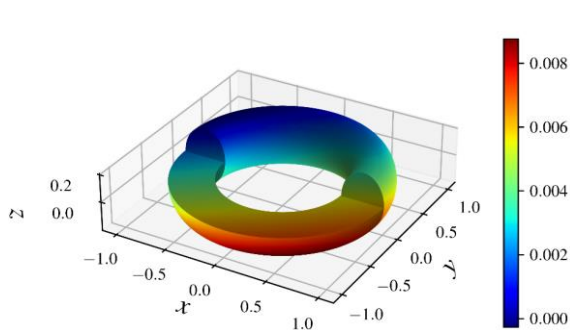

(c) Standard PINN inversion solution

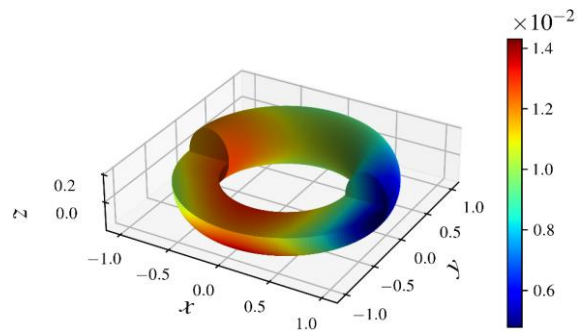

(d) CTL-PINN Absolute error

Fig 10 At $t=195s$, Distribution of source terms functions Q (a) Exact solution; (b) CTL-PINN inversion solution;(c) Standard PINN inversion solution; (d) CTL-PINN Absolute error

## 4.2 Kirchhoff plate dynamic response

The Kirchhoff plate theory is a mathematical model used to determine the stress and deformation of thin plates under the influence of forces and moments (Arf et al., 2023). It is widely applied in engineering to analyze the behavior of thin plates, such as airplane wings, fuselages, and other structural components where the thickness is much smaller than the other dimensions. Its main assumptions are: Mid-surface plane, Straight lines remain straight, Normal lines remain normal, and Constant thickness. Figure 11 shows the schematic of a Kirchhoff rectangular thin plate. The problem of small deflection bending and vibration of the plate is solved based on displacement, with the basic unknown being the deflection of the plate $w$. For plates with varying thickness, the form of the differential equation for plate vibration is as follows:

$$D\nabla^4 w + 2\frac{\partial D}{\partial x}\frac{\partial}{\partial x}\nabla^2 w + 2\frac{\partial D}{\partial y}\frac{\partial}{\partial y}\nabla^2 w + \nabla^2 D \nabla^2 w - (1-\mu)(\frac{\partial^2 D}{\partial x^2}\frac{\partial^2 w}{\partial y^2} - 2\frac{\partial^2 D}{\partial x \partial y}\frac{\partial^2 w}{\partial xy} + \frac{\partial^2 D}{\partial y^2}\frac{\partial^2 w}{\partial x^2}) + \rho h \frac{\partial^2 w}{\partial t^2} = F(x,y,t) \qquad (29)$$

where $(x,y,t) \in [0,a] \times [0,b] \times (0,T]$, $w = w(x,y,t)$ is the deflection function to be determined, $D = \frac{Eh^3}{12(1-\mu^2)}$ is the bending stiffness, $E$ is the modulus of elasticity, $h$ is the thickness of the plate, $\mu$ is the Poisson's ratio, $\rho$ is the density of the plate, and $F(x,y,t)$ is the external load.

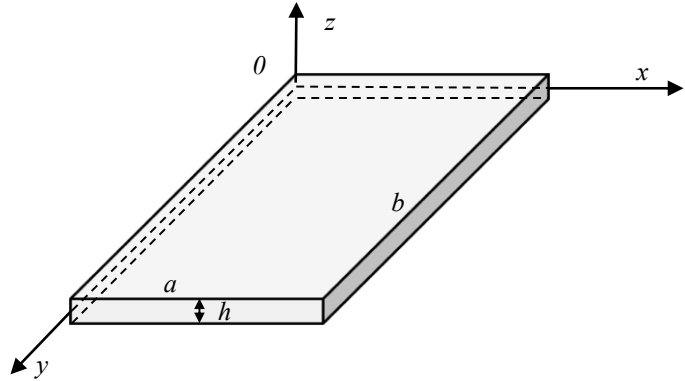

Fig 11 Kirchhoff rectangular sheet diagram

In this study, we set $a = 1\text{m}$, $b = 1\text{m}$, $h = 0.008y + 0.016(\text{m})$, $E = 3 \times 10\text{Pa}$, $\mu = 0.3$, $\rho = 2500\text{kg/m}^3$, and the external load is a uniformly distributed load $(\text{N/m}^2)$. The specific functional form is as follows:

$$F(x,y,t) = 25000(\sin(\frac{t}{2}) + 0.5\cos(\frac{t}{3}) + e^{\frac{-t}{10}}t)\sin(\frac{t}{5}) \qquad (30)$$

With the four edges of the plate fixed, the boundary conditions are:
$$w(0,y,t) = w(1,y,t) = w(x,0,t) = w(x,1,t) = 0 \qquad (31)$$

At the initial moment, the deflection of the plate is zero, hence the initial condition is:
$$w(x, y, 0) = 0 \tag{32}$$

Disregarding the second type of initial conditions, the neural network model is set with 12 hidden layers, each containing 40 neurons. The neural network's input is $(x, y, t)$ and the output is $u_\theta$. The approximate solution for the deflection is set as follows:
$$w_\theta = u_\theta x^2 y^2 (x-1)^2 (y-1)^2 \tag{33}$$

With this setup, $w_\theta$ naturally satisfies the boundary conditions. We only need to consider the initial condition supervised learning loss $\mathcal{L}_i$ and the residual loss $\mathcal{L}_r$. The weights of the loss functions are all set to 1. The number of initial points is $N_i = 500$ and the number of residual points is $N_r = 1200$. These points are obtained using Latin Hypercube Sampling within the corresponding domains. For curriculum learning and transfer learning, the number of additional supervised learning points is $N_{sp} = 1200$.

Figure 12 illustrates the errors of the standard PINN method and the CTL-PINN method across different solution time domains $(0, T]$. Based on the model trained by the standard PINN within $(0, 20s]$, we perform curriculum learning in 4 steps with a step size of 5s, and then transfer learning in 6 steps with a step size of 10s based on the curriculum learning model. Ultimately, the solvable time domain is extended to $(0, 100s]$. It can be observed that, compared to the standard PINN, the CTL-PINN produces more accurate and stable training results, extending the solvable time domain. This demonstrates the effectiveness of the CTL-PINN method in solving the Kirchhoff plate dynamic response, a problem involving higher-order derivatives.

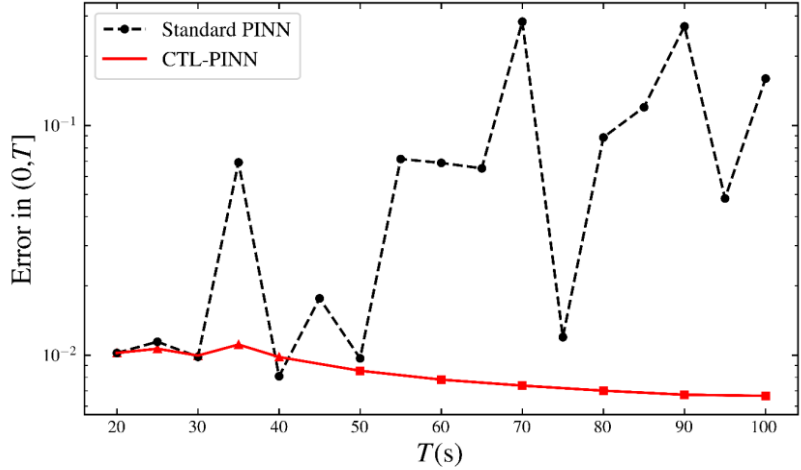

Fig 12 Error variation with respect to the increase of the time domain by standard PINN and CTL-PINN

Figure 13 illustrates the predicted solution and the exact solution at the center of the plate $(0.5, 0.5)$ over the time domain $(0, 100s]$ for standard PINN and CTL-PINN when $T = 100s$. It can be observed that the CTL-PINN predicted solution is in close agreement with the analytical solution, demonstrating the effectiveness of the CTL-PINN method in simulating the long-term Kirchhoff plate dynamic response. On the other hand, the standard PINN predicted solution shows significant deviation from the analytical solution within the interval $(80s, 90s]$ but fits well in other time domains. This also reflects the issue of the standard PINN continuous time model violating the time causality principle.

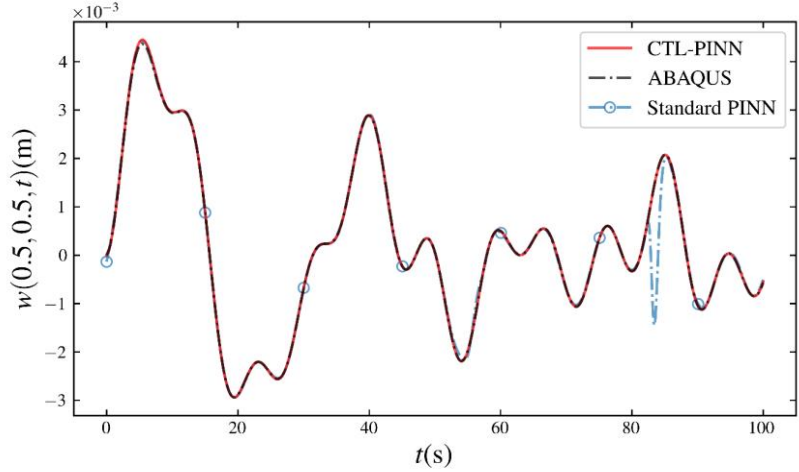

Fig 13  ABAQUS solution, predicted solution by standard PINN and CTL-PINN at point (0.5, 0.5)

Figure 14 depicts the distributions of the ABAQUS solution, CTL-PINN, CTL-PINN Absolute error, standard PINN, and standard PINN Absolute error at $t=100\text{s}$. The difference between CTL-PINN and the ABAQUS solution is small, with an absolute error within $2\times 10^{-3}$, while the standard PINN shows significant deviation from the ABAQUS solution at the center of the plate. The solution results of CTL-PINN are more accurate than those of the standard PINN, demonstrating the superiority of CTL-PINN in simulating the long-term Kirchhoff plate dynamic response.

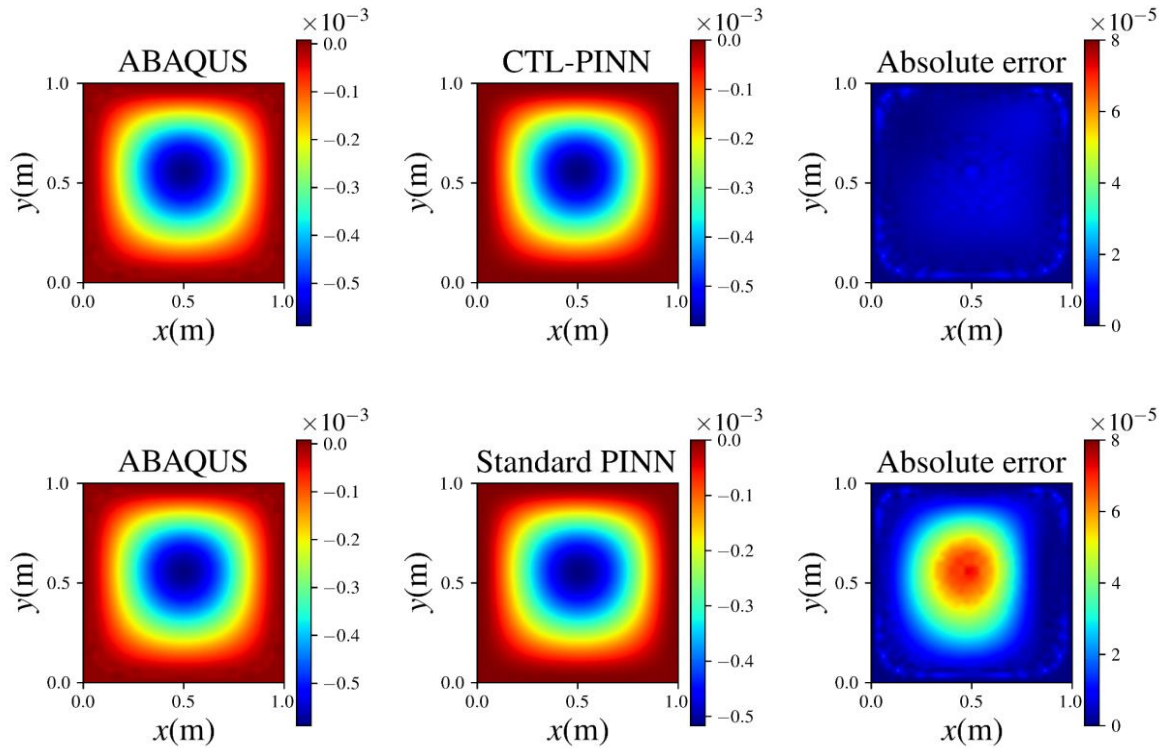

Fig 14  ABAQUS solution, CTL-PINN solution, CTL-PINN Absolute error, standard PINN solution and standard PINN Absolute error at $t = 100s$

## 4.3  Hydrodynamic model of the Three Gorges Reservoir Area

The Three Gorges Reservoir Area (Tang et al., 2019; Q. Zhang & Lou, 2011), located in the central-western region of China, covers a drainage area of 1 million km$^2$ and has an average annual runoff of 451 billion m$^3$. With a normal pool level of 175.0 meters, the reservoir serves multiple functions including flood control, power generation, navigation, and water supply. To provide scientific and rational guidelines for reservoir operation, a hydrodynamic model is developed to simulate and analyze the flow characteristics within the Three Gorges Reservoir Area. Currently, the hydrodynamic model for the Three Gorges Reservoir Area is established based on the Saint-Venant equations (Z. Cao et al., 2006; Islam et al., 2008; Venutelli, 2002; Yi et al., 2017), which are commonly used to describe the one-dimensional flow in open channels:

$$B\frac{\partial Z}{\partial t} + \frac{\partial Q}{\partial x} = q_l \tag{34}$$

$$\frac{\partial Q}{\partial t} + \frac{\partial}{\partial x}(\frac{Q^2}{A}) + gA\frac{\partial Z}{\partial x} + \frac{gn^2|Q|Q}{AR^{4/3}} = q_l U_l \tag{35}$$

Equations (34) and (35) represent the continuity equation and the momentum equation for water flow, respectively. In these equations, $B$ (m) denotes the channel cross-sectional width, $Z$ (m) stands for the water level, $t$ (s) indicates the time, $Q$ (m$^3$/s) is the discharge, $x$ (m) represents the distance along the channel, and $q_l$ (m$^3$/s) refers to the lateral inflow. Additionally, $U_l$ (m/s) is the lateral velocity of inflow, $A$ (m$^2$) signifies the wetted cross-sectional area, and $R$ (m) is the hydraulic radius of the channel section. The gravitational acceleration is denoted by $g$ (m/s$^2$), and $n$ is the roughness coefficient, which ranges from 0.01 to 0.1.

However, during the simulation process, it is essential to determine a critical parameter in the equations—the friction factor, or "roughness coefficient $n$," which directly impacts the accuracy of hydrodynamic simulations. Due to the influence of incoming water and sediment, the roughness coefficient exhibits time-varying characteristics. Its application involves a broad adjustment range, and the calibration and optimization heavily rely on the operator's personal experience. To address this issue, we employed the CTL-PINN method to invert the roughness coefficient based on the measured data, and achieve predictions of discharge and water levels 24 hours in advance.①②

### 4.3.1 River division and data situation

As shown in Figure 15, the river network structure in the Three Gorges area is highly complex due to various factors such as geographical location, topography, and climatic conditions. To accurately simulate the flow dynamics in the main stream of the reservoir, the river network is typically divided into several segments based on its topological structure. Due to data limitations, this study focuses on two segments: Cuntan-Tongluoxia and Changshou-Weidong.

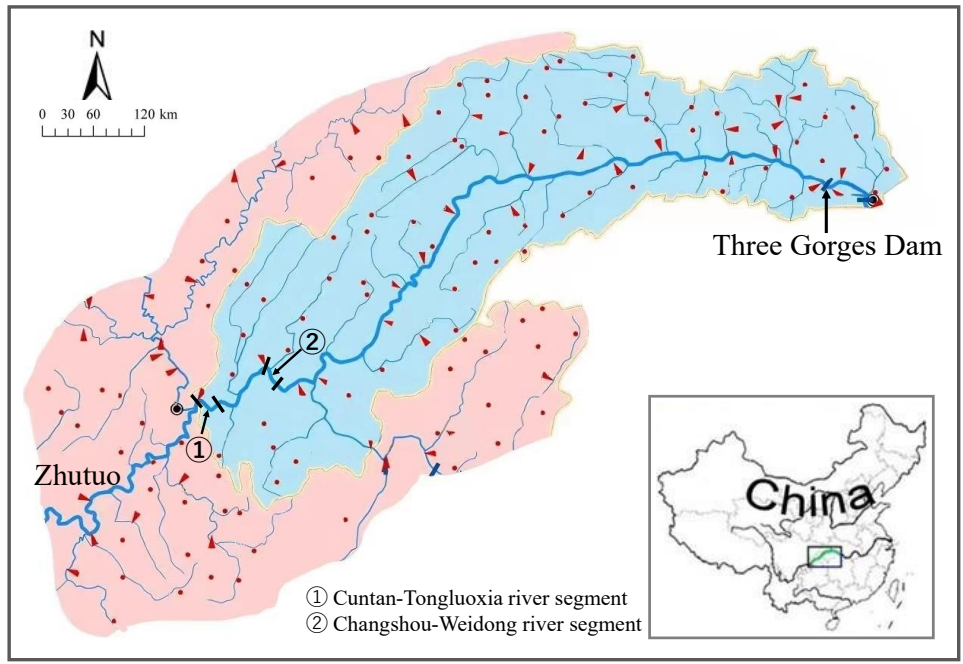

Fig 15 River network structure of the Three Gorges Reservoir Area

The main stream of the reservoir starts at Zhutuo and ends at the Three Gorges Dam, the locations of the Cuntan-Tongluoxia and Changshou-Weidong segments are shown in Figure 15. To determine the change of channel shape in the X-axis direction, a practical and economical approach is to measure cross-sections at regular intervals. In the Cuntan-Tongluoxia segment, three cross-sections (S326.1, S325.1, and S323.1) were measured. In the Changshou-Weidong segment, four cross-sections (S290+1, S290, S289, and S287+2) were measured. The cross-sectional data consists of a series of coordinates $y$ perpendicular to the $x$ direction and the corresponding elevations, representing the river section's topography. Both Cuntan-Tongluoxia and Changshou-Weidong segments have a hydrological station that provides water level data, and discharge data can be obtained at the beginning and end points of the segments, such as Cuntan, Tongluoxia, Changshou, and Weidong. The data situation is summarized in Table 1.

Table 1 Place names and data situation of Three Gorges Reservoir Area

| River segment | Place Name | Data Situation |
| --- | --- | --- |
|  | Cuntan | Discharge |
|  | S326.1 | Cross-section |
|  | Hydrological Station | Water level |
| Cuntan-Tongluoxia | S325.1 | Cross-section |
|  | S323.1 | Cross-section |
|  | Tongluoxia | Discharge |
|  | Changshou | Discharge |
|  | S290+1 | Cross-section |
| Changshou-Weidong | Hydrological Station | Water level |
|  | S290 | Cross-section |
|  | S289 | Cross-section |

| River segment | Place Name | Data Situation |
|---|---|---|
| | S287+2 | Cross-section |
| | Weidong | Discharge |

This section does not have initial and boundary conditions. Discharge data were measured at the beginning and end of the river segments, while water levels were recorded at the hydrological station. The dataset spans from June 1, 2015, 00:00 to June 11, 2015, 13:00, with measurements taken every hour, resulting in a total of 254 data points. The first 230 data points were used for training, and the remaining 24 for testing. Taking June 1, 2015, 00:00 as the reference time point, the time domain range for the training set is $(0, 230\text{h}]$, and the time domain range for the test set is $(230\text{h}, 254\text{h}]$. Table 2 presents the calibration of roughness coefficients for the Three Gorges Reservoir Area in 2015, which can serve as reference values for inverting roughness coefficients. The discharge level in Table 2 typically refer to the average discharge within the relevant time periods. All data in this section were provided by the Science and Technology Research Institute of China Three Gorges Corporation.

Table 2 The calibration results of the roughness coefficient for certain mainstream river segment of the Three Gorges Reservoir Area in 2015.

| Discharge level (m³/s) | 2500 | 3000 | 5000 | 8000 | 10000 | 20000 | 30000 |
|---|---|---|---|---|---|---|---|
| Cuntan-Tongluoxia | 0.0505 | 0.0505 | 0.0405 | 0.0405 | 0.0405 | 0.0405 | 0.0455 |
| Changshou-Weidong | 0.0335 | 0.0335 | 0.0335 | 0.0305 | 0.0305 | 0.0365 | 0.0405 |

4.3.2 Model details

In Equations (34) and (35), the spatial and temporal coordinates $x$ and $t$ are in units of meters (m) and seconds (s), respectively. Directly using these as inputs to the neural network can lead to solution failure due to the large spatial and temporal scales. It is necessary to perform some special processing by first converting the spatial and temporal coordinates $x$ (m) and $t$ (s) into dimensionless forms:

$$x^D = x/1000\text{m}, t^D = t/3600\text{s} \tag{36}$$

where $x^D$ and $t^D$ are dimensionless numbers. This effectively reduces the spatial and temporal resolution of the neural network, making the training of the neural network easier. In fact, we are more concerned with changes on the scale of km and h rather than m and s. Before inputting into the neural network, we further normalize $x^D$ and $t^D$:

$$x^G = \frac{2(x^D - x^D_{\min})}{x^D_{\max} - x^D_{\min}} - 1, t^G = \frac{2(t^D - t^D_{\min})}{t^D_{\max} - t^D_{\min}} - 1 \tag{37}$$

where $x^G$ and $t^G$ range from [-1,1], and they are used as inputs to the neural network. The outputs of the neural network are the dimensionless discharge $Q^D_\theta$ and water depth $H^D_\theta$, where $\theta$ represents the neural network parameters. The neural network in this section has 10 hidden layers, each with 50 neurons. By performing the following calculations, we obtain the discharge $Q_\theta$ (m³/s) and water depth $H_\theta$ (m) in standard units:

$$\begin{cases} Q_\theta = (Q_{\max} - Q_{\min})\text{sigmoid}(Q_\theta^D) + Q_{\min} \\ H_\theta = (H_{\max} - H_{\min})\text{sigmoid}(H_\theta^D) + H_{\min} \end{cases} \quad (38)$$

The range of the sigmoid function is $(0,1)$, and $Q_{\max}, Q_{\min}, H_{\max}$ and $H_{\min}$ are the extrema in standard units. This confines $Q_\theta \in (Q_{\min}, Q_{\max})$ and $H_\theta \in (H_{\min}, H_{\max})$, reducing the difficulty of training the neural network. Based on all cross-sectional data measured in a river section, combined with basic hydraulic knowledge, we can derive the water level $Z_\theta$ (m), hydraulic radius $R_\theta$ (m), wetted cross-sectional area $A_\theta$ (m$^2$), and water surface width $B_\theta$ (m):

$$\begin{cases} Z_\theta = H_\theta + f_{Z\min}^{LI}(x) \\ B_\theta = f_B(H_\theta) \\ A_\theta = f_A(H_\theta) \\ R_\theta = f_R(H_\theta) \end{cases} \quad (39)$$

where $f_{Z\min}^{LI}$, $f_B$, $f_A$, and $f_R$ are derived known functions, with specific derivation processes detailed in the appendix. The final model's inputs $x$ and $t$, and outputs $Z_\theta(x,t)$, $R_\theta(x,t)$, $A_\theta(x,t)$, and $B_\theta(x,t)$, are all in standard units. The loss function integrates the equations with various measured data and is composed of the following parts:

$$\mathcal{L} = \mathcal{L}_{PDE} + \mathcal{L}_{data} = w_{PDE1}\mathcal{L}_{PDE1} + w_{PDE2}\mathcal{L}_{PDE2} + w_{data1}\mathcal{L}_{data1} + w_{data2}\mathcal{L}_{data2} \quad (40)$$

$$\mathcal{L}_{PDE1} = \frac{1}{N_r}\sum_{i=1}^{N_r}\left|B_\theta(x_i,t_i)\frac{\partial Z_\theta(x_i,t_i)}{\partial t} + \frac{\partial Q_\theta(x_i,t_i)}{\partial x} - q_l\right|^2 \quad (41)$$

$$\mathcal{L}_{PDE2} = \frac{1}{N_r}\sum_{i=1}^{N_r}\left|\frac{\partial Q_\theta(x_i,t_i)}{\partial t} + \frac{\partial}{\partial x}\left(\frac{Q_\theta(x_i,t_i)^2}{A_\theta(x_i,t_i)}\right) + gA_\theta(x_i,t_i)\frac{\partial Z_\theta(x_i,t_i)}{\partial x} \right.$$
$$\left. + \frac{gn^2|Q_\theta(x_i,t_i)|Q_\theta(x_i,t_i)}{A_\theta(x_i,t_i)R_\theta(x_i,t_i)^{4/3}} - q_l U_l\right|^2 \quad (42)$$

$$\mathcal{L}_{data1} = \frac{1}{N_Q}\sum_{i=1}^{N_Q}|Q_\theta(x_i,t_i) - Q_{data}(x_i,t_i)|^2 \quad (43)$$

$$\mathcal{L}_{date2} = \frac{1}{N_Z}\sum_{i=1}^{N_Z}|Z_\theta(x_i,t_i) - Z_{data}(x_i,t_i)|^2 \quad (44)$$

where $N_r = 2\times 10^4$ is the number of PDE residual points, obtained using Latin Hypercube Sampling within the corresponding domains. $N_Q = 460$ and $N_Z = 230$ are the number of discharge and water level measurement points, respectively. The position coordinates of the measurement data are in kilometers, and the time coordinates are in seconds. These are converted to standard units of meters and seconds before being input into the model, with various derivatives obtained via automatic differentiation. Since the river segments are divided according to the topological structure of the river, both segments in this study have no major tributaries, and there has been no significant rainfall within the study period. Therefore, the seepage and evaporation of the river segment can be neglected compared to the total discharge, and $q_l$ and $U_l$ are set to 0. The roughness parameter $n \in (0.01, 0.1)$ is an optimizable parameter in the model. The weights of the loss functions, $w_{PDE1}, w_{PDE2}, w_{data1}$ and $w_{data2}$, are adaptively adjusted according to the learning rate annealing algorithm (Wang et al., 2020), which addresses the issue of gradient pathological imbalance. The specific rules are as follows:

$$\begin{cases} w_j^1 = w_{PDE1} = 1 \\ w_j^{i+1} = (1-\alpha)w_j^i + \alpha \frac{1}{w_j^i} \frac{\max(|\nabla_\theta \mathcal{L}_{PDE1}(\theta^i)|)}{\text{mean}(|\nabla_\theta \mathcal{L}_j(\theta^i)|)} \end{cases} \quad (45)$$

where $i$ represents the iteration number, and $j \in \{PDE2, data1, data2\}$. $\nabla_\theta$ denotes the loss gradient with respect to the network parameters. The value of $\alpha$ is set to 0.1. In this study, we stop using this algorithm at an iteration number of $3 \times 10^5$, fixing the values of $w_{PDE1}, w_{PDE2}, w_{data1}$ and $w_{data2}$. Subsequent iterations do not update these values.

For curriculum learning and transfer learning, the roughness parameter $n$ and the neural network parameters $\theta$ inherit the values from the previous training step. The additional supervised learning dataset includes discharge and water level datasets, with the number of additional supervised learning points $N_{sp} = 5000$. The additional supervised learning loss is:

$$\mathcal{L}_{sp} = w_{sp1}\mathcal{L}_{sp1} + w_{sp2}\mathcal{L}_{sp2} \quad (46)$$

where $\mathcal{L}_{sp1}$ and $\mathcal{L}_{sp2}$ represent the supervised learning loss for the discharge dataset and the water level dataset, respectively. The corresponding weights, $w_{sp1}$ and $w_{sp2}$, are also adaptively adjusted using the aforementioned algorithm.

### 4.3.3 Result and discussion

The CTL-PINN method builds on the model trained by the standard PINN method over the time domain $(0, 230h]$, first performing 12 steps of curriculum learning with a step size of 1h, followed by 6 steps of transfer learning with a step size of 2h, thereby extending the solvable time domain to $(0, 254h]$. The standard PINN method directly solves within the time domain $(0, 254h]$, using the measured data from the training set $(0, 230h]$ for both methods. Table 3 records the roughness coefficients, errors of discharge $Q$ and water level $Z$ test sets obtained by the two methods. As shown, the roughness coefficients derived from both methods are generally consistent and close to the reference values. In fact, the inversion of roughness is completed in the first step of the CTL-PINN method, and the roughness value remains nearly unchanged during subsequent curriculum and transfer learning phases. Figure 16 depicts the variation of the loss value and roughness value with respect to the number of iterations during the first step of the inversion process using the CTL-PINN method. It can be observed that both the loss function value and roughness value converge, verifying the reliability of the inversion results.

Table 3 The inverted roughness coefficient results for certain river segments in the Three Gorges Reservoir Area in 2015

| River segment | Average discharge (m³/s) | Roughness reference | Method | Roughness inversion value | Error $Q$ | Error $Z$ |
|---|---|---|---|---|---|---|
| Cuntan-Tongluoxia | 7457 | 0.0405-0.0405 | CTL-PINN | 0.0435 | $1.89 \times 10^{-3}$ | $1.20 \times 10^{-4}$ |
| | | | standard PINN | 0.0442 | $8.22 \times 10^{-2}$ | $1.22 \times 10^{-2}$ |
| Changshou-Weidong | 7812 | 0.0335-0.0305 | CTL-PINN | 0.0316 | $1.88 \times 10^{-3}$ | $4.04 \times 10^{-4}$ |
| | | | standard PINN | 0.0318 | $9.31 \times 10^{-2}$ | $1.16 \times 10^{-2}$ |

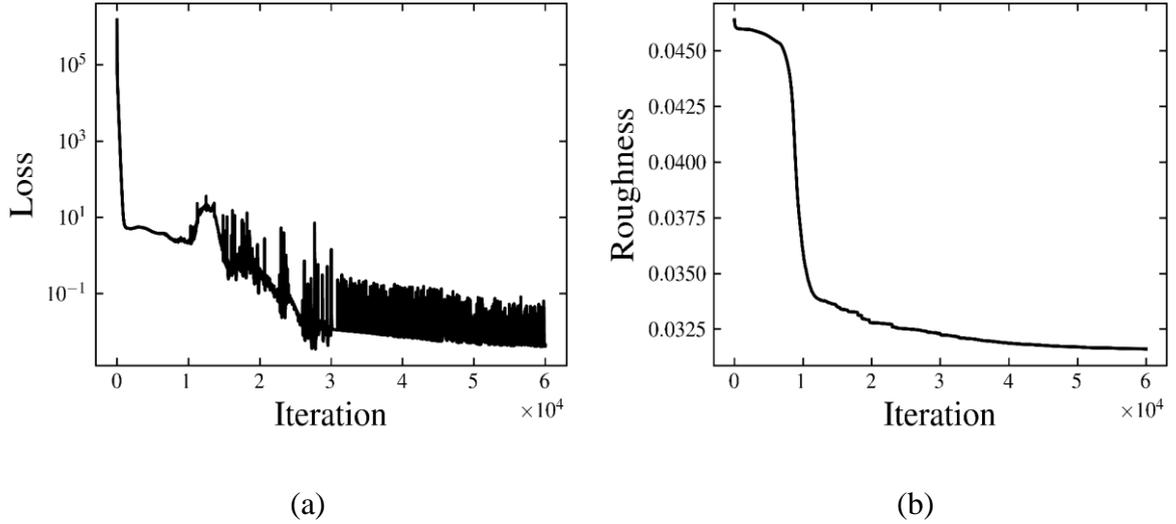

(a)          (b)

Fig 16 The variation of (a) loss value and (b) roughness value with respect to the number of iteration during the first step of the inversion process using the CTL-PINN method in Changshou-Weidong river segment

Comparing the errors of $Q$ and $Z$ test sets, it is evident that the CTL-PINN method significantly outperforms the standard PINN method. Figure 17 illustrates the predicted water level variations at the hydrological station positions within $(0, 254h]$ using the two methods. Within the training set $(0, 230h]$, both the CTL-PINN and standard PINN methods fit the training set well, which is why the roughness values inverted by both methods are generally consistent. However, in the test set $(230h, 254h]$, the CTL-PINN predictions closely match the measured data, while the PINN method shows larger deviations. This demonstrates the superior extrapolation capability of the CTL-PINN method, indicating its potential for real-time prediction of flow and water levels while inverting roughness coefficients.

It is observed that the standard PINN method has high accuracy in the first 2-3 hours of extrapolation prediction, but the prediction accuracy decreases over time. In fact, the standard PINN method only has the constraint of PDEs and no data constraints when predicting values within the prediction time domain (230h,254h]. The high accuracy within the first 2-3 hours is due to the presence of a certain degree of constraint from the data in the training set (0,230h]. However, as the prediction time increases, this constraint effect diminishes. The CTL-PINN method achieves accurate long-term predictions by utilizing this short-term constraint effect from the training set. Both curriculum learning and transfer learning use additional supervised learning data, and the extrapolation step size does not exceed 2 hours. Each time the training set (including measured data and supervised learning data) imposes a good constraint, allowing each extrapolation prediction to proceed.

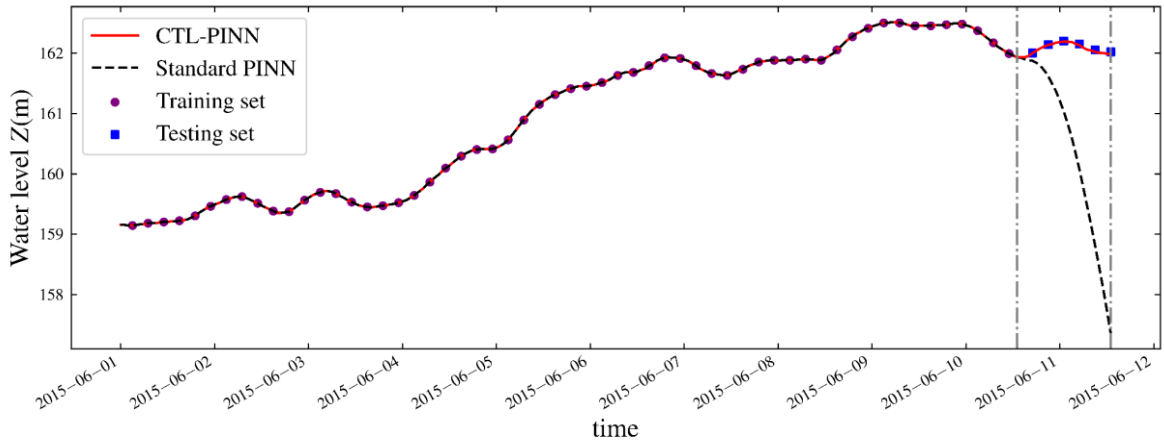

(a) Hydrological station in Cuntan-Tongluoxia river segment

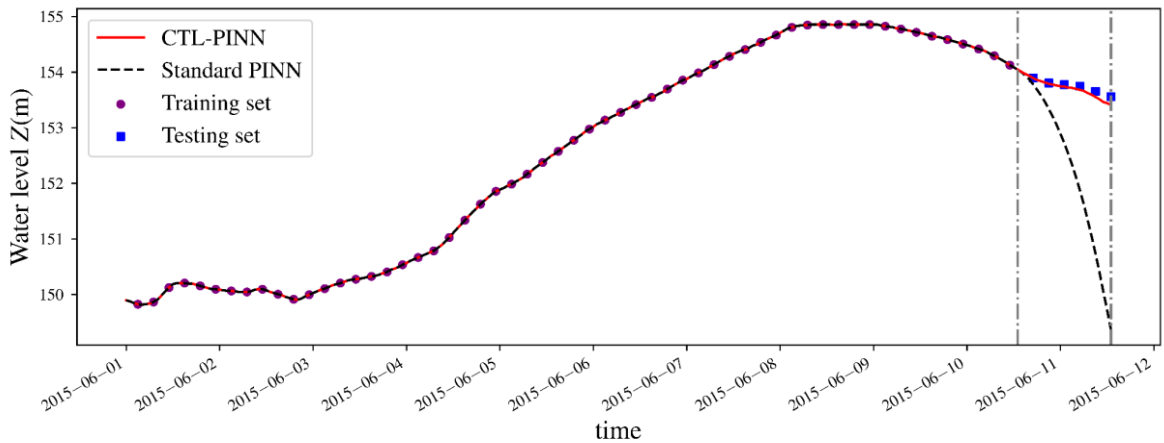

(b) Hydrological station in Changshou-Weidong river segment

Fig 17  The water level variations in the Three Gorges Reservoir Area over time as predicted by CTL-PINN method and standard PINN method

## 5  Conclusion

This paper proposes a curriculum-transfer learning physics-informed neural network, which effectively addresses the issues of poor computational stability and the inability to obtain valid solutions in standard physics-informed neural networks for long-term simulations. By transforming the long-term simulation problem into several short-term simulation subproblems and combining curriculum learning and transfer learning techniques, long-term simulation is achieved. Additionally, this method leverages the neural network parameters and extra supervised learning point information obtained from the current step of the physics-informed neural network training, avoiding the local optimal solution problem in standard physics-informed neural networks. The effectiveness and robustness of this method in simulating long-term problems are demonstrated through several examples, including nonlinear wave propagation, thin plate dynamic response, and the hydrodynamic model of the Three Gorges Reservoir Area. The summary is as follows:

(1) The CTL-PINN method can obtain more accurate solutions than the standard PINN method and can solve for a longer time domain.

(2) Additional supervised learning points can improve the training effect of the neural network. However, since the data for additional supervised learning is predicted by the neural network, there is a certain error between it and the exact values, leading to some error accumulation. Nevertheless, additional measured data can mitigate error accumulation.

(3) In the study of the hydrodynamic model of the Three Gorges Reservoir Area, the variations in water level and discharge in the river segment are not particularly large, and the scenario of heavy rainfall is not considered. Further studies can incorporate rainfall data.

## CRediT authorship contribution statement

**Yuan Guo:** Writing – review & editing, Writing – original draft, Visualization, Validation, Methodology, Investigation, Formal analysis, Data curation, Conceptualization. **Zhuojia Fu:** Writing – review & Editing, Methodology, Supervision, Project administration, Funding acquisition. **Jian Min:** Writing – review & Editing, Methodology, Conceptualization. **Shiyu Lin:** Investigation. **Xiaoting Liu**：Supervision, Funding acquisition, Data curation. **Youssef F. Rashed:** Supervision, Investigation. **Xiaoying Zhuang:** Supervision, Investigation.

## Declaration of competing interest

The authors declare that they have no known competing financial interests or personal relationships that could have appeared to influence the work reported in this paper.

## Data availability

In addition to section 4.3, data and code for other sections will be provided upon request.

## Acknowledgments

The research was supported by the National Natural Science Foundation of China (12122205 and 12372196) and the independent research project of China Three Gorges Corporation (NBZZ202200535).

## APPENDIX

This section introduces the functional relationships between water depth $H$ and water level $Z$, hydraulic radius $R$, wetted cross-sectional area $A$, and water surface width $B$, as well as how to derive specific functions using the measured data from all cross-sections of a river segment. Taking the Changshou-Weidong river segment as an example, it has measured cross-sectional data at sections S326.1, S325.1, and S323.1. Linear interpolation is used to fit the coordinates $y$ and corresponding elevations in the measured data, resulting in three interpolation functions $f_{S326.1}^{LI}(y)$, $f_{S325.1}^{LI}(y)$, and $f_{S323.1}^{LI}(y)$. Linear interpolation is also used to fit the minimum elevation values at sections S326.1, S325.1, and S323.1 with respect to their position coordinates $x$, resulting in the interpolation function $f_{Z\min}^{LI}(x)$. Given the water depth $H(H>0)$, the water level $Z$ can be expressed as:

$$Z = H + f_{Z\min}^{LI}(x) \tag{47}$$

As shown in Figure 18, at a certain section S, let $f_S^{LI}(y) = Z$. Solving this equation yields two solutions, $y_{\min}$ and $y_{\max}$. The water surface width $B$ is:

$$B = y_{\max} - y_{\min} \tag{48}$$

The wetted cross-sectional area $A$ is:

$$A = BZ - \int_{y_{\min}}^{y_{\max}} f_S^{LI}(y)dy \tag{49}$$

The wetted perimeter $P$ is:

$$P = \int_{y_{\min}}^{y_{\max}} \sqrt{1 + \left(\frac{df_S^{LI}(y)}{dy}\right)^2} dy \tag{50}$$

Thus, the hydraulic radius $R$ is:

$$R = \frac{A}{P} \tag{51}$$

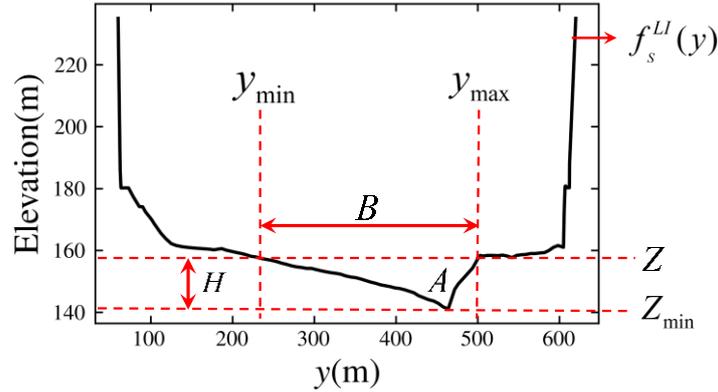

Fig 18 A schematic diagram of the functional relationships between various variables at a certain section S

Taking a series of sufficiently many discrete values within the range of $H$, we substitute them into the above formulas to obtain a series of discrete values for $A$, $B$, and $R$. Using linear interpolation to fit these values, we obtain the following three interpolation functions:

$$\begin{cases} B = f_S^B(H) \\ A = f_S^A(H) \\ R = f_S^R(H) \end{cases} \qquad (52)$$

where $f_S^B$, $f_S^A$, and $f_S^R$ reflect the influence of the topography at a certain section S on the river. In the Changshou-Weidong river segment, it is necessary to comprehensively consider the influence of the topography at all sections on the river. However, in practice, we only have measured data at sections S326.1, S325.1, and S323.1. Therefore, some simplification is required. The following average functions are used to approximate the influence of the topography at all sections on the river:

$$\begin{cases} B = \frac{1}{3}[f_{S326.1}^B(H) + f_{S325.1}^B(H) + f_{S323.1}^B(H)] = f_B(H) \\ A = \frac{1}{3}[f_{S326.1}^A(H) + f_{S325.1}^A(H) + f_{S323.1}^A(H)] = f_A(H) \\ R = \frac{1}{3}[f_{S326.1}^R(H) + f_{S325.1}^R(H) + f_{S323.1}^R(H)] = f_R(H) \end{cases} \qquad (53)$$

where $f_B$, $f_A$, and $f_R$ are known functions derived from the measured data of all cross-sections in the river segment and are applicable to the entire segment. The final specific forms of the functional relationships between water depth $H$, water level $Z$, hydraulic radius $R$, wetted cross-sectional area $A$, and water surface width $B$ are as follows:

$$\begin{cases} Z = H + f_{Z\min}^{LI}(x) \\ B = f_B(H) \\ A = f_A(H) \\ R = f_R(H) \end{cases} \qquad (54)$$

# References

Arf, J., Reichle, M., Klinkel, S., & Simeon, B. (2023). Scaled boundary isogeometric analysis with C1 coupling for Kirchhoff plate theory. *Computer Methods in Applied Mechanics and Engineering*, *415*, 116198.

Beck, C., Becker, S., Grohs, P., Jaafari, N., & Jentzen, A. (2021). Solving the Kolmogorov PDE by means of deep learning. *Journal of Scientific Computing*, *88*, 1–28.

Bengio, Y., Louradour, J., Collobert, R., & Weston, J. (2009). Curriculum learning. *Proceedings of the 26th Annual International Conference on Machine Learning*, 41–48.

Berardi, M., Difonzo, F. V., & Icardi, M. (2025). Inverse Physics-Informed Neural Networks for transport models in porous materials. *Computer Methods in Applied Mechanics and Engineering*, *435*, 117628.

Cao, F., Gao, F., Yuan, D., & Liu, J. (2024). Multistep asymptotic pre-training strategy based on PINNs for solving steep boundary singular perturbation problems. *Computer Methods in Applied Mechanics and Engineering*, *431*, 117222.

Cao, Z., Pender, G., & Carling, P. (2006). Shallow water hydrodynamic models for hyperconcentrated sediment-laden floods over erodible bed. *Advances in Water Resources*, *29*(4), 546–557.

Chen, Z., Lai, S.-K., & Yang, Z. (2024). AT-PINN: Advanced time-marching physics-informed neural network for structural vibration analysis. *Thin-Walled Structures*, *196*, 111423.

Fu, Z., Xu, W., & Liu, S. (2024). Physics-informed kernel function neural networks for solving partial differential equations. *Neural Networks*, *172*, 106098.

Guo, J., Yao, Y., Wang, H., & Gu, T. (2023). Pre-training strategy for solving evolution equations based on physics-informed neural networks. *Journal of Computational Physics*, *489*, 112258.

Han, J., Jentzen, A., & E, W. (2018). Solving high-dimensional partial differential equations using deep learning. *Proceedings of the National Academy of Sciences*, *115*(34), 8505–8510.

Hao, Z., Liu, S., Zhang, Y., Ying, C., Feng, Y., Su, H., & Zhu, J. (2022). Physics-informed machine learning: A survey on problems, methods and applications. *arXiv Preprint arXiv:2211.08064*.

Islam, A., Raghuwanshi, N., & Singh, R. (2008). Development and application of hydraulic simulation model for irrigation canal network. *Journal of Irrigation and Drainage Engineering*, *134*(1), 49–59.

Jagtap, A. D., & Karniadakis, G. E. (2020). Extended physics-informed neural networks (XPINNs): A generalized space-time domain decomposition based deep learning framework for nonlinear partial differential equations. *Communications in Computational Physics*, *28*(5).

Jagtap, A. D., Kharazmi, E., & Karniadakis, G. E. (2020). Conservative physics-informed neural networks on discrete domains for conservation laws: Applications to forward and inverse problems. *Computer Methods in Applied Mechanics and Engineering*, *365*, 113028.

Jeong, H., Batuwatta-Gamage, C., Bai, J., Rathnayaka, C., Zhou, Y., & Gu, Y. (2025). An advanced physics-informed neural network-based framework for nonlinear and complex topology optimization. *Engineering Structures*, *322*, 119194.

Jeong, H., Batuwatta-Gamage, C., Bai, J., Xie, Y. M., Rathnayaka, C., Zhou, Y., & Gu, Y. (2023). A complete physics-informed neural network-based framework for structural topology optimization. *Computer Methods in Applied Mechanics and Engineering*, *417*, 116401.

Karniadakis, G. E., Kevrekidis, I. G., Lu, L., Perdikaris, P., Wang, S., & Yang, L. (2021). Physics-informed machine learning. *Nature Reviews Physics*, *3*(6), 422–440.

Krishnapriyan, A., Gholami, A., Zhe, S., Kirby, R., & Mahoney, M. W. (2021). Characterizing possible failure modes in physics-informed neural networks. *Advances in Neural Information Processing Systems*, *34*, 26548–26560.

Meng, C., Seo, S., Cao, D., Griesemer, S., & Liu, Y. (2022). When physics meets machine learning: A survey of physics-informed machine learning. *arXiv Preprint arXiv:2203.16797*.

Meng, X., Li, Z., Zhang, D., & Karniadakis, G. E. (2020). PPINN: Parareal physics-informed neural network for time-dependent PDEs. *Computer Methods in Applied Mechanics and Engineering*, *370*, 113250.

Penwarden, M., Jagtap, A. D., Zhe, S., Karniadakis, G. E., & Kirby, R. M. (2023). *A unified scalable framework for causal sweeping strategies for Physics-Informed Neural Networks (PINNs) and their temporal decompositions* (arXiv:2302.14227). arXiv.

Qian, Y., Zhu, G., Zhang, Z., Modepalli, S., Zheng, Y., Zheng, X., Frydman, G., & Li, H. (2024). Coagulo-Net: Enhancing the mathematical modeling of blood coagulation using physics-informed neural networks. *Neural Networks*, *180*, 106732.

Quan, L., Qian, F., Liu, X., & Gong, X. (2016). A nonlinear acoustic metamaterial: Realization of a backwards-traveling second-harmonic sound wave. *The Journal of the Acoustical Society of America*, *139*(6), 3373–3385.

Rabczuk, T., Song, J.-H., Zhuang, X., & Anitescu, C. (2019). *Extended finite element and meshfree methods*. Academic Press.

Raissi, M., Perdikaris, P., & Karniadakis, G. E. (2019). Physics-informed neural networks: A deep learning framework for solving forward and inverse problems involving nonlinear partial differential equations. *Journal of Computational Physics*, *378*, 686–707.

Roy, A., Nehra, R., Langrock, C., Fejer, M., & Marandi, A. (2023). Non-equilibrium spectral phase transitions in coupled nonlinear optical resonators. *Nature Physics*, *19*(3), 427–434.

Shukla, K., Jagtap, A. D., & Karniadakis, G. E. (2021). Parallel physics-informed neural networks via domain decomposition. *Journal of Computational Physics*, *447*, 110683.

Sirignano, J., & Spiliopoulos, K. (2018). DGM: A deep learning algorithm for solving partial differential equations. *Journal of Computational Physics*, *375*, 1339–1364.

Tang, H., Wasowski, J., & Juang, C. H. (2019). Geohazards in the three Gorges Reservoir Area, China – Lessons learned from decades of research. *Engineering Geology*, *261*, 105267.

Tasaketh, I., Cen, W., Liu, B., & Zhai, Y. (2022). Nonlinear Seismic Response of SV Wave Incident Angle and Direction to Concrete Gravity Dam. *International Journal of Civil Engineering*, *20*(12), 1479–1494.

Thomas, J. W. (2013). *Numerical partial differential equations: Finite difference methods* (Vol. 22). Springer Science & Business Media.

Thuerey, N., Holl, P., Mueller, M., Schnell, P., Trost, F., & Um, K. (2021). Physics-based deep learning. *arXiv Preprint arXiv:2109.05237*.

Venutelli, M. (2002). Stability and accuracy of weighted four-point implicit finite difference schemes for open channel flow. *Journal of Hydraulic Engineering*, *128*(3), 281–288.

Wang, S., Teng, Y., & Perdikaris, P. (2020). *Understanding and mitigating gradient pathologies in physics-informed neural networks* (arXiv:2001.04536). arXiv.

Weiss, K., Khoshgoftaar, T. M., & Wang, D. (2016). A survey of transfer learning. *Journal of Big Data*, *3*(1), 9.

Xiao, S., Bai, J., Jeong, H., Alzubaidi, L., & Gu, Y. (2025). A meshless Runge-Kutta-based Physics-Informed Neural Network framework for structural vibration analysis. *Engineering Analysis with Boundary Elements*, *170*, 106054.

Yi, Y., Tang, C., Yang, Z., Zhang, S., & Zhang, C. (2017). A One-Dimensional Hydrodynamic and Water Quality Model for a Water Transfer Project with Multihydraulic Structures. *Mathematical Problems in Engineering*, *2017*(1), 2656191.

Younas, U., Muhammad, J., Nasreen, N., Khan, A., & Abdeljawad, T. (2024). On the comparative analysis for the fractional solitary wave profiles to the recently developed nonlinear system. *Ain Shams Engineering Journal*, *15*(10), 102971.

Zhang, E., Dao, M., Karniadakis, G. E., & Suresh, S. (2022). Analyses of internal structures and defects in materials using physics-informed neural networks. *Science Advances*, *8*(7), eabk0644.

Zhang, Q., & Lou, Z. (2011). The environmental changes and mitigation actions in the Three Gorges Reservoir region, China. *Environmental Science & Policy*, *14*(8), 1132–1138.

Zhao, W., Lei, M., & Hon, Y.-C. (2022). An improved finite integration method for nonlocal nonlinear Schrödinger equations. *Computers & Mathematics with Applications*, *113*, 24–33.

Zheng, P.-B., Zhang, Z.-H., Zhang, H.-S., & Zhao, X.-Y. (2023). Numerical Simulation of Nonlinear Wave Propagation from Deep to Shallow Water. *Journal of Marine Science and Engineering*, *11*(5), 1003.

Zienkiewicz, O. C., Taylor, R. L., Nithiarasu, P., & Zhu, J. (1977). *The finite element method* (Vol. 3). Elsevier.